\documentclass{article}

     \PassOptionsToPackage{numbers, compress}{natbib}
\usepackage[main, final]{neurips_2026}
\usepackage[utf8]{inputenc} % allow utf-8 input
\usepackage[T1]{fontenc}    % use 8-bit T1 fonts
\usepackage{url}            % simple URL typesetting
\usepackage{booktabs}       % professional-quality tables
\usepackage{amsfonts}       % blackboard math symbols
\usepackage{nicefrac}       % compact symbols for 1/2, etc.
\usepackage{microtype}      % microtypography
\usepackage{xcolor}         % colors
\usepackage{graphicx}
\usepackage{subcaption}
\usepackage{booktabs} % for professional tables
\usepackage{hyperref}

\newcommand{\quotett}[1]{\texttt{\detokenize{#1}}}
\newcommand{\blocktag}[2]{%
  \tcbox[
    on line,
    arc=0pt,
    boxrule=0pt,
    left=1pt,right=1pt,top=0.5pt,bottom=0.5pt,
    colback=#1,
    nobeforeafter,
    tcbox raise base
  ]{#2}%
}

\usepackage{amsmath}
\usepackage{amssymb}
\usepackage{mathtools}
\usepackage{amsthm}
\usepackage{algorithm}
\usepackage{algorithmic}
\usepackage{float}
\definecolor{highlight}{RGB}{255,50,0}   
\definecolor{continuation}{RGB}{100,130,255} 
\definecolor{branch}{RGB}{255,10,220}  
\definecolor{branch_2}{RGB}{0,200,60}     
\definecolor{boxbg}{RGB}{252,252,252}
\definecolor{nodeedge}{RGB}{60, 60, 60}
\definecolor{successgreen}{RGB}{46, 125, 50}
\definecolor{failred}{RGB}{198, 40, 40}
\definecolor{expandorange}{RGB}{220, 20, 60} 
\definecolor{selectionblue}{RGB}{0, 0, 205}  
\definecolor{strongred}{RGB}{255, 0, 0}
\definecolor{strongred}{RGB}{255, 0, 0} 
\definecolor{stepblue}{RGB}{0,102,204}
\definecolor{stepgreen}{RGB}{0,153,0}
\definecolor{steporange}{RGB}{230,140,0}
\definecolor{steppurple}{RGB}{128,0,180}

\newenvironment{highlightblock}[1]{%
  \begin{tcolorbox}[
    enhanced,
    breakable,
    colback=#1,
    colframe=#1,
    boxrule=0pt,
    arc=1pt,
    left=1pt,
    right=1pt,
    top=1pt,
    bottom=1pt,
    boxsep=0pt,
    before skip=0pt,
    after skip=0pt,
    width=\linewidth
  ]
  \ttfamily\hyphenchar\font=-1
}{%
  \end{tcolorbox}
}

\usepackage[utf8]{inputenc}
\usepackage{tikz}
\usepackage{amssymb}
\usepackage{xcolor}
\usepackage{pifont}
\usepackage{caption}
\usepackage{multirow}
\usepackage{graphicx}
\definecolor{grayrow}{gray}{0.9}
\usepackage[most]{tcolorbox}
\usetikzlibrary{shapes.arrows, arrows.meta, positioning, fit, backgrounds, shapes.geometric}
\newcommand{\highlighttoken}[2]{%
    \setlength{\fboxsep}{0.5pt}
    \colorbox{#1}{\texttt{\strut #2}}%
}
\usepackage[table]{xcolor}

\usepackage{xspace}

\newcommand{\ours}{BFS-PO\xspace}

\usepackage[capitalise]{cleveref}  
\crefname{section}{sec.}{sec.}
\crefname{equation}{eq.}{eq.}
\crefname{figure}{fig.}{fig.}
\crefname{table}{tab.}{tab.}
\crefname{appendix}{app.}{app.}
\crefname{algorithm}{alg.}{alg.}

\newcommand{\algcomment}[1]{\hfill$\triangleright$ #1}

\usepackage{pifont}
\newcommand{\cmark}{\text{\ding{51}}\xspace}
\newcommand{\xmark}{\text{\ding{55}}\xspace}

\theoremstyle{plain}

\theoremstyle{definition}

\theoremstyle{remark}

\usepackage[textsize=tiny]{todonotes}

\title{BFS-PO: Best-First Search for Large Reasoning Models}

\author{
  \textbf{Fiorenzo Parascandolo\textsuperscript{1,2}\thanks{
  Emails: fiorenzo.parascandalo@unimore.it;
  enver.sangineto@unimore.it} , Wenhui Tan\textsuperscript{3}, Enver Sangineto\textsuperscript{1}\footnotemark[1] ,  Xiaoyi Yu\textsuperscript{3},} \\ \textbf{Ruihua Song\textsuperscript{3}, Rita Cucchiara\textsuperscript{1}}  \\
  \textsuperscript{1}  University of Modena and Reggio Emilia, 
  Modena,  Italy, \\
  \textsuperscript{2} University of Pisa, Pisa, Italy \\
  \textsuperscript{3} Gaoling School of Artificial Intelligence, 
  Renmin University of China, 
  Beijing, China \\
}

\begin{document}

\maketitle

\begin{abstract}
 Large Reasoning Models (LRMs)   have shown excellent performance
in reasoning tasks using long Chains of Thought (CoTs).
 However, this has also led to a significant increase
of computational costs and the generation of  verbose output, a phenomenon known as {\em overthinking}.
The tendency to overthink is often exacerbated by Reinforcement Learning (RL) algorithms such as GRPO/DAPO. In this paper, we propose \ours, an RL algorithm which alleviates this problem  using a Best-First Search
exploration strategy.
  Specifically,
the reasoning chains generated at training time by \ours are  organized into a search tree, within which the shortest correct sequence is selected as the best and expanded using a backtracking mechanism based on maximum entropy nodes.  
In this way, we bias the exploration of the solution space towards the search for increasingly shorter solutions, training the LRM to generate progressively more concise answers.
Using different benchmarks and base LRMs, we show that \ours can simultaneously increase the LRM accuracy and shorten its reasoning chains. Our code and our trained models are publicly available at: \url{https://github.com/aimagelab/BFS-PO}. 
\end{abstract}

\section{Introduction}
\label{sec.Introduction}

Large Reasoning Models (LRMs) have recently achieved impressive accuracy results using inference-time scaling
\cite{OpenAI-o1,DeepSeek-R1}, in which the
reasoning process is decomposed into a  (long) sequence of simpler steps,
called  Chain of Thought (CoT) \cite{CoT}.
However, inference-time
scaling usually leads to very long reasoning chains, which often consist of 
thousands or even tens of thousands of tokens  \cite{ballon2025relationshipreasoningperformancelarge,ThinkPrune},
producing high computational and memory overhead. This phenomenon is called {\em overthinking}, and it is due to the presence of redundant steps \cite{rpc}, errors in reasoning \cite{fatemi2025concise} and the use of many words for the sole purpose of textual coherence \cite{Coconut}.
Moreover, several recent studies have empirically demonstrated that 
 {\em excessive CoT length  and accuracy are strongly negatively correlated} \cite{fatemi2025concise,underthinkingoverthink,optimalscaling,dontoverthink,finishsearch,shorterbetter}.
 For instance, in DAPO \cite{DAPO} and in \cite{fatemi2025concise,liu2025}, the authors  observe that Reinforcement Learning (RL) algorithms like GRPO \cite{GRPO} 
encourage longer responses when the {\em advantage} is negative, due to the CoT length-based normalization contained in the objective function
(more details in App. \ref{app.GRPO}).

A  common method to reduce overthinking is based on Supervised Fine-Tuning (SFT) of a pre-trained LRM using a dataset containing a short ground truth CoT associated with each question
(\Cref{sec.Related}). One of the pioneering approaches of this paradigm is TokenSkip \cite{tokenskip}, which utilizes GPT-4 to associate each token in a given CoT with an importance score. Less important tokens are pruned, and the LRM is fine-tuned with the resulting dataset. 
Many other works \cite{A-star,verbosity,limopro,dualprocess,verithinker,self,token-budget,C3oT,Break,ConCISE,Extra-CoT}
follow this direction and are based on the creation of a dataset of triplets  (question, CoT, answer), where the correct answer to a question is associated with a short CoT. 
However, training an LRM using  SFT  is often suboptimal \cite{DeepSeek-R1}, especially with a limited dataset, and can lead to overfitting. Empirical results
show that most of these works succeed in reducing
the length of the CoTs, but at the cost of also reducing the accuracy.

In this paper, we follow a different direction: instead of using a static dataset of pre-computed short CoTs for SFT, we propose an RL algorithm in which   {\em exploration is biased toward correct, short reasoning paths}.
In this way, it is the LRM itself that  generates its dataset of concise CoTs while it is trained.
Specifically, we propose Best-First Search Policy Optimization
(\ours), an RL algorithm in which the exploration process is guided by a  Best-First Search (BFS) strategy on a search tree.
Unlike GRPO-type algorithms, where multiple answers to a given question are generated independently of each other, in \ours, sampling is conditioned on the best solution found so far, where ``best'' is defined as the {\em shortest correct} solution.
Importantly, \ours does not require external modules to evaluate partial solutions and decide which ones to expand. Indeed, tree search-based strategies have been recently adopted in other works \cite{treeThoughts,LLMguidedtree,11123142}. However, these methods  need a Process Reward Model (PRM) \cite{ProcessRewardModel}, i.e. a module 
which can assess the quality of the current reasoning step and guide exploration within the search tree. Unfortunately, a PRM is
usually implemented as an additional Large Language Model (LLM), which needs to be trained with supervised datasets containing annotations for partial solutions. 
Moreover, using a PRM requires additional computational resources during training or inference.
In contrast,  we avoid the need for a PRM by evaluating only complete solutions. Specifically, once a  full solution is generated, we use a standard answer verifier \cite{lambert2025tulu} to assess its correctness. Then, the solution length is used to select the shortest correct reasoning chain. 
Once the current best solution is selected, we use a backtracking mechanism which  interprets this solution as a path in a search tree, and it identifies the most promising forking node where the  full solution should be split in order to pursue a different path.  Since we do not have an external PRM, we select the backtracking node using the generation uncertainty.
Indeed, as  empirically shown in different recent works 
\cite{beyond-80-20,SwiReasoning,deepthinkconfidence,reasoningexplorationentropyperspective},  
the highest entropy CoT tokens   behave as {\em forking tokens} in LRM reasoning, and they correspond to {\em those points in which the LRM takes the most important decisions}.
Intuitively, if the entropy in predicting the next token is low, it means that the token is being used by the LLM primarily to ensure linguistic coherence; conversely, if the entropy is high, it indicates a branching point, as the LLM perceives that it could take different paths to continue its reasoning \cite{beyond-80-20}.
Hence,
we use the highest entropy token as the backtracking point, and  the partial path from the question to this point
to bias the exploration of alternative partial paths (\Cref{fig.tree}). 

Recently, Hou et al.  
 proposed  TreeRL \cite{treerl}, in which, similarly to \ours, a search tree is constructed using the token generation uncertainty for branching. However, unlike our entropy-based selection,   TreeRL uses the 
  negative log-likelihood of the next token as the criterion to select (multiple)
forking tokens in a given path. 
Moreover, in TreeRL,
the goal of the tree search is to efficiently replace  classic Monte Carlo Tree Search (MCTS) 
\cite{MCTS}
methods, and it is not based on a specific search strategy like Best First. 
In fact, in \cite{treerl} the $N$ nodes with maximum uncertainty are expanded in parallel (where $N$ is an empirically set hyperparameter) with the aim of estimating the {\em value} of a node by simulating a Monte Carlo method.
Finally,  the advantage proposed in \cite{treerl} 
 compares the node value with the value of the
root and the value of its parent node. In contrast,
we expand only the correct, shortest solution found so far, with the aim of implementing a greedy search in the discrete space of solutions that leads to a local optimum, and we compute the advantage in GRPO/DAPO style.

We evaluate \ours  using different reasoning benchmarks and base LRMs, and we compare it with both SFT and RL-based baselines. Unlike most existing literature on CoT shortening,
where brevity is achieved at the cost of an accuracy drop with respect to the zero-shot model,
 \ours can simultaneously reduce the average CoT length {\em and} increase the accuracy, 
 both compared to the zero-shot model, and also compared to an RL fine-tuning using strong baselines such as DAPO.

\begin{figure*}[ht]
    \centering
    
    \resizebox{0.7\textwidth}{!}{
\begin{tikzpicture}[scale=1.2, transform shape] 
    \tikzset{
        modern_tree/.style={
            every node/.style={
                circle, draw=black, fill=gray!5, 
                minimum size=0.9cm, 
                thick, inner sep=0pt, 
                font=\sffamily\small, align=center 
            },
            edge from parent/.style={
                draw=black, ultra thick, -{Stealth[length=6pt]}, 
                shorten >=1pt, shorten <=1pt
            }, 
            level 1/.style={sibling distance=1.5cm, level distance=1.5cm},
            level 2/.style={sibling distance=1.5cm, level distance=1.5cm},
            level 3/.style={sibling distance=1.4cm, level distance=1.5cm},
            level 4/.style={sibling distance=1.3cm, level distance=1.5cm},
            level 5/.style={sibling distance=1.2cm, level distance=1.5cm}
        },
        dot_node/.style={draw=none, fill=none, minimum size=0.5cm},
        context_ellipse/.style={ellipse, draw=black!50, dashed, fill=gray!10, inner sep=6pt, line width=1.2pt},
        pure_text/.style={draw=none, fill=none, rectangle, font=\sffamily\large},
        sel_node/.style={draw=blue, fill=blue!10, line width=1.5pt},
        sel_edge/.style={edge from parent/.style={draw=blue, line width=2.5pt, -{Stealth[length=7pt]}}},
        black_edge/.style={edge from parent/.style={draw=black, ultra thick, -{Stealth[length=6pt]}}},
        expansion_path/.style={
            edge from parent/.style={draw=red, ultra thick, -{Stealth[length=6pt]}},
            every node/.append style={draw=red, line width=1.2pt}
        },
        % Simboli OK/X ridimensionati a 2.5
        success/.style={label={below:\textcolor{green}{\scalebox{2.5}{\textbf{\checkmark}}}}},
        fail/.style={label={below:\textcolor{red}{\scalebox{2.5}{\textbf{\texttimes}}}}},
        prog_label_text/.style={font=\sffamily\Huge\bfseries, color=black},
        title_style/.style={font=\sffamily\Huge\bfseries, color=black!70},
        sel_title_style/.style={font=\sffamily\Huge\bfseries, color=blue!70},
        weight_label/.style={draw=none, fill=none, font=\sffamily\Huge, xshift=15pt},
        leaf_label/.style={draw=none, fill=none, font=\sffamily\Huge, xshift=15pt},
        q_arrow/.style={draw=black, ultra thick, -{Stealth[length=6pt]}},
        phase_label/.style={font=\sffamily\Huge\bfseries, color=black},
        transition_arrow/.style={
            draw=black, 
            fill=black, 
            shape=single arrow, 
            single arrow head extend=3pt, 
            minimum height=2.2cm, 
            minimum width=1.1cm,
            inner sep=2pt
        }
    }

    % --- (a) SAMPLING ---
    \node (treeA) at (0, 0.5) [anchor=north] {
        \begin{tikzpicture}[modern_tree]
            \node (n1_A) at (0,0) {};
            \node[dot_node] (dots_A) at (0,-1.5) {\Large $\vdots$};
            \node (n3_A) at (0,-3.0) {}
                child { node {} child { node {} child { node {} child { node [fail] {} } } } }
                child { node {} child { node {} child { node {} child { node [fail] {} } } } }
                child { node {} child { node {} child { node {} child { node {} child { node {} child { node [success] {} } } } } } } 
                child { node {} child { node {} child { node {} child { node {} child { node [success] {} } } } } };
            \draw[q_arrow] (n1_A) -- (dots_A);
            \draw[q_arrow] (dots_A) -- (n3_A);
            \begin{scope}[on background layer]
                \node[context_ellipse, fit=(n1_A) (dots_A) (n3_A)] (ellA) {};
            \end{scope}
            \node[pure_text, right=10pt of ellA, font=\sffamily\itshape\Huge] {q};
        \end{tikzpicture}
    };
    \node (alignerA) at (0, -18.5) {}; 
    \node[prog_label_text] (labelA) at (0, -22.5) {(a)}; 

    % --- (b) PATH SELECTION ---
    \node (treeB) at (10.0, 0.5) [anchor=north] {
        \begin{tikzpicture}[modern_tree]
            \node (n1_B) at (0,0) {};
            \node[dot_node] (dots_B) at (0,-1.5) {\Large $\vdots$};
            \node (n3_B) at (0,-3.0) {}
                child [black_edge] { node {} child { node {} child { node {} child { node [fail] {} } } } }
                child [black_edge] { node {} child { node {} child { node {} child { node [fail] {} } } } }
                child [black_edge] { node {} child { node {} child { node {} child { node {} child { node {} child { node [success] {} } } } } } } 
                child [sel_edge] { node [sel_node, label={[weight_label]right:$w^b$}] {} 
                    child [sel_edge] { node [sel_node] {} 
                        child [sel_edge] { node [sel_node] {} 
                            child [sel_edge] { node [sel_node, label={[leaf_label]right:$o_{j}$}] {} child { node [sel_node, success] {} } } 
                        } 
                    } 
                };
            \draw[q_arrow] (n1_B) -- (dots_B);
            \draw[q_arrow] (dots_B) -- (n3_B);
            \begin{scope}[on background layer]
                \node[context_ellipse, fit=(n1_B) (dots_B) (n3_B)] (ellB) {};
            \end{scope}
            \node[pure_text, right=10pt of ellB, font=\sffamily\itshape\Huge] {q};
        \end{tikzpicture}
    };
    \node[phase_label] (phaseB) at (10.0, -18.5) {Selection};
    \node[prog_label_text] (labelB) at (10.0, -22.5) {(b)}; 

    % --- (c) ITERATIVE GROWTH ---
    \node (treeC) at (21.5, 0.5) [anchor=north] { 
        \begin{tikzpicture}[modern_tree]
            \node (n1_C) at (0,0) {};
            \node[dot_node] (dots_C) at (0,-1.5) {\Large $\vdots$};
            \node (n3_C) at (0,-3.0) {}
                child { node {} child { node {} child { node {} child { node [fail] {} } } } }
                child { node {} child { node {} child { node {} child { node [fail] {} } } } }
                child { node {} child { node {} child { node {} child { node {} child { node {} child { node [success] {} } } } } } } 
                child { node [xshift=3.5cm, label={[weight_label]right:$w^b$}] {} [sibling distance=1.4cm]
                    child [expansion_path] { node {} child { node {} child { node [success] {} } } }
                    child [expansion_path] { node {} child { node {} child { node {} child { node [success] {} } } } }
                    child { node {} child { node {} child { node {} child { node [success] {} } } } }
                    child [expansion_path] { node {} child { node {} child { node {} child { node {} child { node [success] {} } } } } }
                    child [expansion_path] { node {} child { node {} child { node [fail] {} } } }
                };
            \draw[q_arrow] (n1_C) -- (dots_C);
            \draw[q_arrow] (dots_C) -- (n3_C);
            \begin{scope}[on background layer]
                \node[context_ellipse, fit=(n1_C) (dots_C) (n3_C)] (ellC) {};
            \end{scope}
            \node[pure_text, right=10pt of ellC, font=\sffamily\itshape\Huge] {q};
        \end{tikzpicture}
    };
    \node[phase_label] (phaseC) at (21.5, -18.5) {Expansion};
    \node[prog_label_text] (labelC) at (21.5, -22.5) {(c)}; 

    % --- (d) FINAL SELECTION ---
    \node (treeD) at (34.5, 0.5) [anchor=north] { 
        \begin{tikzpicture}[modern_tree]
            \node (n1_D) at (0,0) {};
            \node[dot_node] (dots_D) at (0,-1.5) {\Large $\vdots$};
            \node (n3_D) at (0,-3.0) {}
                child [black_edge] { node {} child { node {} child { node {} child { node [fail] {} } } } } 
                child [black_edge] { node {} child { node {} child { node {} child { node [fail] {} } } } } 
                child [black_edge] { node {} child { node {} child { node {} child { node {} child { node {} child { node [success] {} } } } } } } 
                child [sel_edge] { node [sel_node, xshift=3.5cm, label={[weight_label]right:$w^b$}] {} [sibling distance=1.4cm]
                    child [sel_edge] { node [sel_node] {} child [sel_edge] { node [sel_node] {} child { node [sel_node, success] {} } } } 
                    child [black_edge] { node {} child { node {} child { node {} child { node [success] {} } } } } 
                    child [black_edge] { node {} child { node {} child { node {} child { node [success] {} } } } } 
                    child [black_edge] { node {} child { node {} child { node {} child { node {} child { node [success] {} } } } } } 
                    child [black_edge] { node {} child { node {} child { node [fail] {} } } } 
                };
            \draw[q_arrow] (n1_D) -- (dots_D);
            \draw[q_arrow] (dots_D) -- (n3_D);
            \begin{scope}[on background layer]
                \node[context_ellipse, fit=(n1_D) (dots_D) (n3_D)] (ellD) {};
            \end{scope}
            \node[pure_text, right=10pt of ellD, font=\sffamily\itshape\Huge] {q};
        \end{tikzpicture}
    };
    \node[phase_label] (phaseD) at (34.5, -18.5) {Selection};
    \node[prog_label_text] (labelD) at (34.5, -22.5) {(d)}; 

    \node (final_dots) [pure_text, anchor=west] at (42.0, -8.0) {\Huge \textbf{\dots}};

    % --- Frecce NERE ---
    \node[transition_arrow] at (15.75, -18.5) {};
    \node[transition_arrow] at (28, -18.5) {};

    % --- BOX E TITOLI ---
    \begin{scope}[on background layer]
        \node[draw=black!20, dashed, ultra thick, rounded corners, fill=gray!5, fit=(treeA) (alignerA), inner sep=35pt] (dapoBox) {};
        \node[title_style, anchor=south, yshift=20pt] at (dapoBox.north) {GRPO/DAPO};
        
        \node[draw=blue!40, ultra thick, rounded corners, fill=blue!5, fit=(treeB) (treeD) (phaseB) (phaseD) (final_dots), inner sep=35pt] (ourBox) {};
        \node[sel_title_style, anchor=south, yshift=20pt] at (ourBox.north) {BFS-PO};
    \end{scope}
\end{tikzpicture}

    }
     \caption{A schematic comparison between GRPO/DAPO (a) and \ours (b-d). In (a), the sampling mechanism of GRPO/DAPO is represented as a simple tree with only one forking node (the root node, conditioned on the question $q$). 
    \cmark and \xmark represent correct and incorrect answers, respectively.
    In (b) and (d), \textcolor{blue}{the current best solution} is selected, while in (c)  \textcolor{red}{new branches} are added to the tree. For simplicity, in this figure we use $G=4$.}
    \label{fig.tree}
 \end{figure*}

\section{Related work}
\label{sec.Related}

Due to its significant practical relevance, many works have recently been published on overthinking reduction. In this section, we briefly review the main methods, organized  into different categories.

{\bf Budget forcing} approaches \cite{s1simpletesttimescaling,thinkless,token-budget,efficientlyscaling,SyncThink,lee2025how}
for CoT shortening are based on specifying a maximum  number
of thinking tokens  (usually via a specific prompt) and forcing an early exit by adding an end-of-thought token once this number has been generated. The advantage of budget forcing is that it can be used as a training-free approach. However, its early exit usually leads to an accuracy drop with respect to the zero-shot base model.

{\bf SFT with short CoTs.}
As mentioned in \Cref{sec.Introduction}, one of the most popular CoT shortening paradigm is based on constructing a dataset with ground truth short CoTs associated with each question, and then using SFT to train the model. Approaches in this category differ mainly in how the dataset is built and how to assess the importance of each token, e.g., using an auxiliary LLM
\cite{tokenskip,dualprocess,C3oT,Extra-CoT},  the influence of a token on the response likelihood \cite{A-star,verbosity,limopro,ConCISE} or other methods \cite{verithinker,self,token-budget,Break,Extra-CoT}. In contrast, we do not explicitly build a training dataset containing concise CoTs, but we let the LRM itself generate short CoTs, conditioning the exploration strategy toward the shortest correct solution.

{\bf RL with length penalty.} Another common CoT shortening paradigm is based on RL algorithms which include a length penalty in their reward \cite{sel2025llms,dualprocess,ThinkPrune,Extra-CoT}. 
For instance, in \cite{shorterbetter}, the penalty is computed with respect to the current shortest positive answer, while in \cite{training-efficiently,O1-Pruner}, it is compared to the average length of the sampled  CoTs.  
In \cite{fang2026thinkless,Extra-CoT}, a previous SFT phase with a compression control token is used to teach the model when to shorten the reasoning.
We adopt the length penalty of DAPO \cite{DAPO} %(\Cref{sec.Preliminaries}) 
and we compare with DAPO in \Cref{sec.Experiments}, showing that \ours generates shorter CoTs with comparable or even slightly higher accuracy.

{\bf Steering vector.}
Ma et al. \cite{CoT-Valve}
build  a dataset that associates CoTs of different lengths with each question. They then train LoRA models \cite{lora} with both long
and short CoTs. The difference in parameter space between the two models indicates the direction to shorten the CoTs. A similar approach is used in \cite{controlling}, where the steering vector is computed in the activation space of some LRM layers rather than in the parameter space, and using PCA instead of parameter arithmetic.

{\bf Compression} methods \cite{anchorbased,rpc,LightThinker} compress the tokens of the generated CoTs, e.g., using specific attention masks which force the LRM to store information in intermediate tokens while discarding the others. Usually, this compression results in a decrease in the final accuracy.

{\bf Latent CoTs.} The idea behind this category of methods
\cite{implicit,explicitcotimplicit,Coconut,Looped,codi,token-assorted,scaling-up,Variational,Shortcuts,colar,Albert}
is that human reasoning is often abstract and not always verbal, thus, rather than generating CoTs composed of discrete tokens, the LRM can generate continuous vectors in a latent space. 
Relaxing the textual coherence constraint, a sequence of latent tokens can potentially be shorter than a textual CoT. Although appealing, approaches belonging to this class have so far lagged behind the accuracy achieved by LRMs based on textual CoTs.
One of the reasons is likely that sampling from a continuous distribution during RL training is more difficult than sampling from a distribution over a vocabulary of discrete tokens \cite{lcm}.  Soft Thinking \cite{soft-think} partially alleviates this problem using the LRM discrete distribution to softly aggregate all the vocabulary tokens. However, very recent works 
\cite{demystifying-soft,softtokenshardtruths}
have shown that vanilla implementations of Soft Thinking often underperform their discrete counterparts, and its sampling mechanism mostly boils down to greedy decoding.

{\bf LRM with tree search.} Methods using a tree search in conjunction with an LRM usually do not aim to shorten the generated CoTs, but to extend the linear reasoning, typical of a CoT, into a structured search. The latter includes, for example, techniques such as Breadth-First Search  or Depth-First Search  \cite{11123142}. 
For instance, Tree of Thoughts \cite{treeThoughts,LLMguidedtree} 
break the reasoning process into atomic thoughts (where, e.g., a thought is a sentence), organized in a tree.
Then, they 
use auxiliary LLMs and specific prompts to evaluate the current tree frontier and expand its nodes (i.e., thoughts). This mechanism has been extended in \cite{besta2024got} to direct graphs. Hou et al.
\cite{treerl} observe that MCTS \cite{MCTS}
cannot be easily applied to LRMs because of its high computational costs (see \Cref{sec.Introduction}). Thus, they propose TreeRL, in which, similarly to \ours, 
 the exploration is organized using a  tree search. 
However, they do not use BFS
  to explore the tree nor entropy to measure uncertainty,
and they compute the value of the internal nodes of the  tree by cumulating the rewards on its leaves, in this way replacing the random and costly exploration in a MCTS with an efficient search. %based on the next-token uncertainty 
We compare \ours with TreeRL in \Cref{sec.Experiments}, showing that our method drastically outperforms TreeRL.

% Moreover, in addition to the differences mentioned in \Cref{sec.Introduction}, in TreeRL, non-leaf tokens
% belonging to multiple paths are repeatedly included in the
% optimization process, increasing the risk of overfitting on these
% partial sequences \cite{treerl}. Instead, we propose a different formulation of the advantage that relies only on the
% current subtree and allows shared paths to be computed only once, resulting in a much faster algorithm 
% and with a higher stability of the supervision signal (see \Cref{sec.efficiency,sec.Experiments}).

\section{Preliminaries}
\label{sec.Preliminaries}

{\bf Problem Setup.}
Let  $\pi_{\theta}$ be a policy model
 (i.e. a base LRM) with parameters $\theta$, and ${\cal D} = \{ (q_j, a_j^*) \}$ the training dataset, 
 composed of problems $(q_j, a_j^*)$,
 where $q_j$ is the question and $a_j^*$ the corresponding ground truth answer. Given $q_j$, the model autoregressively generates a token sequence $o_j = (w_1, ..., w_t, ..., w_{|o_j|})$ according to $o_j \sim \pi_{\theta}(\cdot|q_j)$, where $w_t \in V$ is a textual token and $V$ the vocabulary of all tokens. We indicate with $o_{j,t} = w_t$ the t-th token of the reasoning chain $o_j$
 (we will use $o_{j,t}$ and $w_t$ interchangeably). The output $o_j$
 is often formatted using special tokens (e.g., $\texttt{<think>}$, $\texttt{</think>}$) to delimit the intermediate reasoning chain  from the final answer $a_j$. A verifier $v(\cdot)$, usually based on simple parsing and string matching techniques,  extracts $a_j$ from $o_j$ and compares it with $a_j^*$: $v(o_j,a_j^*) = 1$ if $a_j = a_j^*$, and 0 otherwise.
The goal is to  fine-tune $\pi_{\theta}$ to decrease the average length  of its reasoning chains 
while keeping (or increasing) its  accuracy. 

{\bf GRPO and DAPO.}
The success of DeepSeek-R1 \cite{DeepSeek-R1} has popularized GRPO \cite{GRPO}, an RL algorithm which has been shown to greatly increase the accuracy of LRMs. The main idea in GRPO is to replace the costly  {\em value model} in PPO \cite{PPO} with an {\em advantage} computed with respect to the mean  reward of a group of $G$ different outputs $\{o_i\}_{i=1}^G$ corresponding to the same question $q$.
%(more details in \Cref{app.GRPO}).
Specifically, given $(q,a^*) \in {\cal D}$, $\pi_{\theta_{old}}$ is used to sample $\{o_i\}_{i=1}^G$, and, for each $o_i$, 
the advantage is computed using:

\begin{minipage}{\linewidth}
\begin{equation}
\label{eq.advantage-GRPO}
\hat{A}_{i,t} = \frac{R_i - \operatorname{mean}(\{ R_i \}_{i=1}^G)}{\operatorname{std}(\{ R_i \}_{i=1}^G)},
\end{equation}
\end{minipage}

where $R_i$ is the sample-level reward 
%($R_i = 1$ if $v(o_i,a^*) = 1$ and $R_i = -1$ if $v(o_i,a^*) = 0$) 
and $\hat{A}_{i,t}$ is the token-level advantage, same for 
all  tokens $o_{i,t} \in o_i$.
DAPO \cite{DAPO} 
is an improved version of GRPO (see App. \ref{app.GRPO}), in which the same advantage is used in the following objective:

\begin{equation}
\label{eq.DAPO}
\begin{split}
{\cal J}_{DAPO} (\theta) = &  \mathbb{E}_{(q,a^*) \sim {\cal D}, \{o_i\}_{i=1}^G \sim \pi_{\theta_{old}}(\cdot|q)} \\
& \left[ \frac{1}{\sum_{i=1}^G |o_i|} \sum_{i=1}^G \sum_{t=1}^{|o_i|} 
\left( 
min (r_{i,t}(\theta) \hat{A}_{i,t}, clip(r_{i,t}(\theta),  
  1 -  \epsilon_l, 1 + \epsilon_h) \hat{A}_{i,t} 
\right) 
\right] \\
& \textnormal{s.t.:} \: \: 0 < |\{ o_i | v(o_i,a^*) = 1 \}| <  G,
\textnormal{where:} \: \:
r_{i,t}(\theta) = \frac{\pi_{\theta}(o_{i,t}|q, o_{i,<t})}{\pi_{\theta_{old}}(o_{i,t}|q, o_{i,<t})}.
\end{split}
\end{equation}

\noindent
The constraint $0 < |\{ o_i | v(o_i,a^*) = 1 \}| <  G$ 
is used to filter out trivial questions (where all the answers are correct) and questions which are too difficult for the current policy model $\pi_{\theta}$, both leading to a 0 advantage. When this constraint is violated, $q$ is discarded and a new problem $(q,a^*)$ is sampled from ${\cal D}$. 
Moreover, in DAPO   a penalty for very long sequences is added to the reward $R_i$:

 \begin{equation}
 \label{eq.OverlongPunishment}
{\cal R}_{len}(o) = 
\begin{cases}
    0 & \text{$|o| \leq L_{max} - L_{cache}$}\\
    \frac{(L_{max} - L_{cache}) - |o|}{L_{cache}} & \text{$L_{max} - L_{cache} < |o| \leq L_{max}$}\\
    -1 & \text{$L_{max} < |o|$,}\\
  \end{cases}
\end{equation}
 
\noindent
and the final reward is computed using:

 \begin{equation}
 \label{eq.DAPO-tot-reward}
R_i = 
\begin{cases}
    1 + {\cal R}_{len}(o_i) & \text{$v(o_i,a^*) = 1$}\\
    -1 + {\cal R}_{len}(o_i) & \text{$v(o_i,a^*) = 0$.}\\
  \end{cases}
\end{equation}

In this paper, we adopt the objective function of DAPO, including its length penalty, and, for a fair comparison, we {\em use the same  hyperparameter values of DAPO} (more details in  App. \ref{app.GRPO}). 
In \Cref{fig.tree} (a) we schematically show the group of $G$ outputs, which can be thought of as independent paths of a simple tree containing only one branching node (the root node).

\section{Method}
\label{sec.Method}

The $G$ different outputs used in GRPO and DAPO (see \Cref{sec.Preliminaries} and \Cref{fig.tree}) are generated independently of one another and are used to explore the solution space, providing potentially different solutions for the same question  $q$.
However, classical optimization methods for planning and problem-solving usually organize the exploration of a discrete state-space of solutions  using a search tree and strategies such as BFS, while employing mechanisms like backtracking to re-examine and modify a partial solution \cite{russel2010}. 
For instance, in our case, the benefit of a BFS strategy over multiple independent random sampling is that we can steer  the exploration towards a local optimum of the solution space by progressively refining 
the current best solution to make it increasingly shorter.
To implement this idea, we first select the best solution generated so far, where ``best'', in our case, means correct and shortest.
Specifically, if ${\cal T} = \{o_i \}_{i=1}^G$, then let $o_j \in {\cal T}$ be such that:

\begin{equation}
\label{eq.best-solution}
v(o_j,a^*) = 1, \: \:
|o_j| \leq min_{o_i \in {\cal T}}|o_i|.
\end{equation}

Note that, since we adopt the resampling strategy of DAPO, i.e., we resample $(q,a^*)$ if $|\{ o_i | v(o_i,a^*) = 1 \}| =  0$ or $|\{ o_i | v(o_i,a^*) = 1 \}| =  G$ (\Cref{sec.Preliminaries}), then  there always exists at least one $o_j$ such that $v(o_j,a^*) = 1$.
For a given $q$, $o_j$, selected using \Cref{eq.best-solution},
is the best solution generated so far. We now want to condition the exploration process using $o_j$. More specifically, we want to explore the solution space in the neighbourhood of $o_j$, searching for a shorter, correct solution.
To do so, we need a backtracking mechanism, i.e., if we interpret $o_j$ as a path in a search tree (${\cal T}$), 
we need to identify the most promising forking node where the reasoning process should follow a different path.  
As mentioned in \Cref{sec.Introduction}, and following the empirical findings in \cite{beyond-80-20}, we use the highest entropy tokens for backtracking.
Specifically, we select the {\em backtracking} token $w^b \in o_j$ (\Cref{fig.tree} (b)) using:

\begin{equation}
\label{eq.backtrack-token}
w^b = o_{j,t^*} =   \arg \max_{o_{j,t} \in o_j} H(o_{j,t}),
\end{equation}
where:
\begin{equation}
\label{eq.backtrack-token-entropy}
H(o_{j,t}) = - \sum_{n=1}^{|V|} \pi_{\theta}(w_n|q, o_{j,\leq t}),
log(\pi_{\theta}(w_n|q, o_{j,\leq t})).
\end{equation}

Using $o_{j,t^*}$, we expand the current tree by adding  new branches. 
In more detail, we use $(o_{j,1}, ..., o_{j,t^*})$ as a suffix which is concatenated to the question $q$ to condition the generations of $G$ new outputs, sampled using $\{o_i\}_{i=1}^G \sim \pi_{\theta_{old}}(\cdot|q, o_{j,1}, ..., o_{j,t^*})$. The new partial solutions 
$\{o_i\}_{i=1}^G$ are added to the search tree ${\cal T}$
(\Cref{fig.tree} (c))  and the process is iterated.
The full algorithm is shown in \Cref{alg.Tree}, where $B$ % = \{ w_1^b, w_2^b, ... \}$ 
is the set of backtracking tokens. At  iteration $k$, the new best
solution $o_j$ is selected (line 5) from all complete paths from the root to a leaf using \Cref{eq.best-solution}. 
Then, a backtracking token ($o_{j,t^*}$) is chosen (line 6) in the current shortest correct solution $o_j$, 
where we exclude from the selection the previously chosen backtracking tokens ($o_{j,t^*} \notin B$) to increase the exploration diversity.
To avoid an exponential tree growth, we use a fixed number of expansion iterations, $K$ ($K = 3$, see \Cref{sec.Ablation}), which plays a role analogous to that of $G$ in GRPO/DAPO, limiting the exploration of the solution space for the sake of efficiency.

\subsection{Optimization}
\label{sec.Optimization}

Once ${\cal T}$ is built,
we associate a reward and an advantage to each token
which are then used to compute the objective function.
%we extract all the complete reasoning chains, which correspond to all the paths in ${\cal T}$ from the root to a leaf.
Specifically, if $w_n$ is the token associated to a generic node in ${\cal T}$,  we indicate with $w_m = P(w_n)$ the parent of $w_n$.
Similarly, we denote with $C(w_n) = \{w_{m_1}, ..., w_{m_k} \}$  the set of  children of $w_n$.  
If $w_n \in B$, then 
 $k= G+1$; for leaf nodes, $k=0$; and   $k=1$ for all the other nodes.
Moreover, %if $w_n$ is a leaf,
from each leaf node $w_i$,
we use $P(\cdot)$ to extract a full reasoning path recursively  
until we reach the root ($q$), obtaining the output sequence $o_i = (w_1, ..., w_i) = (P(...P(w_i)...), ..., P(P(w_i)), P(w_i), w_i)$.
Note that, while ${\cal T}$ is
only partially ordered, each $o_i$ is totally ordered; it  is therefore consistent with the standard causal attention mechanism of a decoder-only transformer.
If ${\cal T}$ has $N$ leaves,  we extract $N$ complete reasoning chains $CoTs = \{ o_i \}_{i=1}^N$. 
For each $o_i = (w_1, ..., w_i) \in CoTs$,
we use \Cref{eq.DAPO-tot-reward} to compute 
 a {\bf reward} $R_i$, which is assigned to  the {\em leaf} $w_i$.
The rewards for the internal nodes are recursively computed  averaging the rewards of their children. Specifically, if $w_n$ is the token of a non-leaf node, its reward $R(w_n)$ is:
\begin{equation}
    \label{eq.recursive-reward}
    R(w_n) = \frac{1}{|C(w_n)|} \sum_{w_m \in C(w_n)} R(w_m).
\end{equation}
Note that, if $w_n \notin B$, then $R(w_n) = R(C(w_n))$.
This recursive reward computation is similar to the ``value backpropagation'' in MCTS \cite{MCTS}, and it is different from the
sample-level reward of GRPO/DAPO, in which all the tokens of a CoT receive the same reward (\Cref{sec.Preliminaries}). 
The benefits of our  token-level reward approach include both a partial reduction of the credit assignment problem and a decrease in the variance of the estimated values, thanks to the  averaging in \Cref{eq.recursive-reward}.

After a reward has been assigned to each token, for each $o_i \in CoTs$ and each $o_{i,t} \in o_i$, we compute the {\bf advantage}, using:

\begin{minipage}{\linewidth}
\begin{equation}
\label{eq.advantage-BFS-PO-global}
\hat{A}_{i,t} = \frac{R(o_{i,t}) - \operatorname{mean}(\{R_j | o_j \in CoTs \})}{\operatorname{std}(\{R_j | o_j \in CoTs \})}.
\end{equation}
\end{minipage}

Finally, our surrogate {\bf objective function} is:
\begin{equation}
\label{eq.our-loss}
\begin{split}
{\cal J}_{\ours} (\theta) = &  \mathbb{E}_{(q,a^*) \sim {\cal D},
{\cal T} \sim \mu_{BFS}(\cdot|q; \pi_{\theta_{old}}) } \\
& \left[ \frac{1}{\sum_{i=1}^N |o_i|} \sum_{i=1}^N \sum_{t=1}^{|o_i|} 
\left( 
min (r_{i,t}(\theta) \hat{A}_{i,t}, clip(r_{i,t}(\theta),  
 1 -  \epsilon_l, 1 + \epsilon_h) \hat{A}_{i,t} 
\right) 
\right], \\
\end{split}
\end{equation}
where the outer sum iterates over the $N$ sequences of $CoTs = \{ o_i \}_{i=1}^N$  extracted from  ${\cal T}$, and 
$\mu_{BFS}(\cdot|q; \pi_{\theta_{old}})$
represents the entire sampling procedure, which depends on both the old policy
$\pi_{\theta_{old}}$ and \Cref{alg.Tree}.
However, since the dependence of $\mu_{BFS}(\cdot|q; \pi_{\theta_{old}})$ 
 on the procedure described in \Cref{alg.Tree} makes the estimation of this likelihood non straightforward,  we simplify the importance sampling ratio in \Cref{eq.our-loss} using only $\pi_{\theta_{old}}$, and we set: 
$r_{i,t}(\theta) = \frac{\pi_{\theta}(o_{i,t}|q, o_{i,<t})}{\pi_{\theta_{old}}(o_{i,t}|q, o_{i,<t})}$.
% Note that, using $\pi_{\theta_{old}}$ instead of $\mu_{BFS}$ in the denominator of 
% $r_{i,t}(\theta)$,
%   \Cref{eq.our-loss} becomes a surrogate objective function, rather than an unbiased estimator of the standard DAPO objective function.

% \textnormal{and:} \: \: CoTs = \{ o_i \}_{i=1}^N  \: \: \textnormal{is extracted from}  \:  {\cal T}.
% \end{split}
% \end{equation}

At {\bf inference time} we use standard, single-chain, autoregressive sampling (without tree): given a question $q$, we sample $o \sim \pi_{\theta}(\cdot|q)$.

\begin{center}
\begin{minipage}{\linewidth}
\begin{algorithm}[H]
  \caption{Tree Construction}
  \label{alg.Tree}
  \begin{algorithmic}[1]
    \STATE {\bfseries Input:} sample problem $(q,a^*) \sim {\cal D}$,
        policy model $\pi_{\theta_{old}}$, number of expansion iterations $K$.
    %\STATE[] $\texttt{/* Initialization */}$ %\algcomment{Initialization}
    \STATE ${\cal T} = \{o_i \}_{i=1}^G \sim \pi_{\theta_{old}}(\cdot|q)$ s.t. $0 < |\{ o_i | v(o_i,a^*) = 1 \}| <  G$
    \algcomment{{\footnotesize Initialization}}
    \STATE $B = \{ \empty \}$   
    \FOR{$k=1$ {\bfseries to} $K$}
        %\STATE[] $\texttt{/* Selection */}$ %\algcomment{Selection}
        \STATE Select a complete path $o_j$ using \Cref{eq.best-solution}  
        \algcomment{{\footnotesize Selection}}   
        \STATE $o_{j,t^*} =   \arg \max_{o_{j,t} \in o_j, o_{j,t} \notin B} H(o_{j,t})$ 
        %\STATE $\texttt{/* Expansion */}$
        %\STATE[] $\texttt{/* Expansion */}$ %\algcomment{Expansion}
        \STATE ${\cal N}= \{o_i \}_{i=1}^G \sim \pi_{\theta_{old}}(\cdot|q, o_{j,1}, ..., o_{j,t^*})$ \algcomment{{\footnotesize Expansion}}
        \STATE ${\cal T} \leftarrow {\cal T} \cup {\cal N}$
        \STATE $B \leftarrow B \cup \{ o_{j,t^*} \}$
    \ENDFOR
  \end{algorithmic}
\end{algorithm}
\end{minipage}
\end{center}

\subsection{Implementation details}
\label{sec.Discussion}

{\bf Computational cost.}
Most of the computational cost in both DAPO and \ours comes from generating the reasoning chains, while the optimization step is the same and has the same computational cost for both methods (\Cref{sec.Optimization}) and the overhead of tree construction is negligible. To make the comparison fair, we use  the same number of  generated sequences per question  and the same number of questions  per generation batch. Specifically, for each question $q$, let $G_D$ denote the number of output sequences generated by DAPO, and  $G_B$ denote the number of outputs generated by \ours at each expansion step $k$ ($k = 1, ..., K$, see \Cref{alg.Tree}). Since \ours performs $K + 1$ tree expansions (including the initial generation at step 2), it generates $N = (K+1)G_B$ sentences for each question (where $N = |CoTs|$, see \Cref{sec.Optimization}). Thus, we set $G_D = N$, which implies $G_B = G_D/(K+1)$ (we use $G_D = 32$ in all our experiments).
Moreover, for both methods we use the same number of questions $Q$ per generation batch.
However, while in DAPO the $QG_D$ sentences are generated in parallel, in \ours we need $K+1$ sequential steps to generate the same number $QN$ of outputs.
Nevertheless, even though \ours requires  $K$ more generation steps than DAPO to generate the same number of sentences, the overall  wall-clock time is comparable for two reasons.
 First, after the initial expansion in step 2, each generation in \ours starts from a backtracking point, 
 thus the tokens from the root to the backtracking point have already been generated and do not need to be recomputed. 
 Second, as shown in \Cref{fig.training_curves_gsm8k} (App. \ref{app.TrainingCurves}), during training, \ours reduces the average length of the generated sequences much faster than DAPO, which leads to a large computational advantage. In all our experiments we use the same fixed training time budget for all the methods, and, as shown in App. \ref{app.TrainingCurves}, in this period \ours and DAPO complete almost the same number of training steps.
 To further validate this observation, 
 in \Cref{app.Training-time-comparison}
 we compare  the  wall-clock training time of \ours with its closest competitors, DAPO and TreeRL, 
 showing that there is no significant difference between the three approaches.

{\bf Discussion.}
Although we recursively compute a {\em token-level} reward  (\Cref{eq.recursive-reward}), the advantage is based
on  a {\em sample-level} mean and standard deviation, using all the elements in  $CoTs$ (\Cref{eq.advantage-BFS-PO-global}).
In our preliminary experiments, we extended the local reward computation  to the advantage as well, by replacing the mean and standard deviation in \Cref{eq.advantage-BFS-PO-global} with values calculated  at each backtracking node. However, our empirical results show that the difference is not significant. More importantly, this forking-node based  advantage has a computational issue  when, for instance, all $k$ branches of $w_n$ yield a positive reward, which can result in a token-level advantage of zero. Although this issue could be resolved by adopting a resampling mechanism—similar to the one used in DAPO (\Cref{sec.Preliminaries})—{\em at each backtracking node}, doing so significantly increases computation time.

As mentioned above, in \ours  some  output sequences share part of their tokens with other sequences because they have common sub-paths from the root to a backtracking node (see \Cref{fig.tree}).
Hence, since  on average \ours observes fewer tokens in a batch than DAPO, the question
is: does this lead to a decrease in the diversity of solutions
explored during training? 
In \Cref{sec.Experiments} we show that \ours, in most experiments, achieves an accuracy equal to or higher than that of DAPO (as well as shorter responses),  which empirically shows that \ours can generalize as well as or better than DAPO. We believe that the main reason for this is that \ours generates solutions that differ more in their ``reasoning logic'', compared to DAPO. 
In fact,  as shown in \cite{beyond-80-20}, 
the tokens with low  entropy are mainly used by an LRM to ensure the linguistic coherence of the sentence, while the
highest entropy tokens are the actual decision points which determine the logical trajectory of 
the generated solution. Thus, the textual difference of the independently generated chains in DAPO does not necessarily correspond to a deep semantic difference. In contrast, at each expansion step 
of \ours
(\Cref{alg.Tree}), we
branch  
 the highest entropy node of the current best solution,  
and in this way we allow the LRM to explore different trajectories by making decisions ``logically'' different
(\Cref{sec.Ablation}).

\section{Experiments}
\label{sec.Experiments}

\subsection{Experimental setup}
\label{sec.ExpSetup}

We adopt two experimental protocols taken from TokenSkip \cite{tokenskip} and Extra-CoT \cite{Extra-CoT}.
 Extra-CoT \cite{Extra-CoT} is a recently proposed work which combines SFT with short CoTs and RL with a length penalty (\Cref{sec.Related}), where the variant in which RL is used is called CHRPO.
Specifically, we use the In-Domain ({\bf ID}) experimental setup taken from
TokenSkip \cite{tokenskip} and adopted also in Extra-CoT \cite{Extra-CoT}, in which training is done on the training split of  GSM8K and testing on the testing split of the same dataset.  In this case, following 
\cite{tokenskip}, the {\em  base models} are:
Qwen2.5-3B-Instruct,
Qwen2.5-7B-Instruct, 
Qwen2.5-14B-Instruct \cite{Qwen},
Llama3.2-3BInstruct
and LLaMA-3.1-8B-Instruct
\cite{Llama}, and the  {\em token budget} is 512 tokens. 
Moreover, we use  the experimental setup proposed in Extra-CoT \cite{Extra-CoT}, which is based on Out-of-Domain ({\bf OOD}) training and testing, where the
 {\em training} phase uses   a subset of 60k problem samples randomly extracted from 
MetaMathQA-395K \cite{MetaMath}\footnote{In \cite{Extra-CoT}, the authors use an additional dataset, CAMEL \cite{camel}, for training their Compressor and additional data from MetaMathQA in the RL phase, which we do not use for simplicity.}.
On the other hand,  {\em testing} is based on different  {\em benchmarks}:  
GSM8K \cite{GSM8K},  MATH-500 \cite{lightman2024lets},  AMC2023 \cite{AI-MO-Team} 
and 
 MMLU-STEM \cite{MMLU}.
GSM8K, MATH-500 and  AMC2023 contain mathematical problems, while MMLU-STEM is based on
Science, Technology, Engineering, and Mathematics (STEM) subjects.
Following \cite{Extra-CoT}, in the OOD setup we use Qwen3-1.7B \cite{yang2025qwen3technicalreport}
as the {\em  base model}  
with a {\em token budget} of 4096 tokens.

{\bf Metrics.}
We use the evaluation
script adopted in \cite{Extra-CoT}  for answer extraction and correctness verification.  Besides
the standard {\em Accuracy}, used to measure the answer
correctness,  we use {\em Average Length} to report the average number of tokens in the  reasoning chains.
Moreover, since a drastic reduction in CoT length usually also results in a reduction in accuracy (\Cref{sec.Introduction,sec.Related}), 
we use the {\em Accuracy Efficiency Score (AES)}, proposed in \cite{O1-Pruner},
which combines the two metrics
 and evaluates the trade-off between
improving accuracy and reducing computational costs (see App. \ref{app.AES}  for more details).

{\bf Baselines.}
We compare \ours with different methods representing most of the paradigms mentioned in \Cref{sec.Related}: 1) Zero-shot
and standard SFT;
2) Training-free budget forcing methods, including
length-control prompts (LC-Prompt \cite{lee2025how}) and hard truncation of the output sequence (Truncation \cite{lee2025how}) ; 
3) SFT with short CoTs, including TokenSkip \cite{tokenskip} and Extra-CoT \cite{Extra-CoT}, both with different compression ratios; 4) 
RL with length penalty, including Thinkless (DeGRPO) \cite{fang2026thinkless} and Extra-CoT with RL (CHRPO) \cite{Extra-CoT};
5) DAPO, which is a state-of-the-art RL method for general-purpose LRM training; and
6) LRM with tree search, including Tree of Thought \cite{treeThoughts} and TreeRL \cite{treerl}.

\subsection{Ablation}
\label{sec.Ablation}

In our ablation experiments we use the ID  scenario (\Cref{sec.ExpSetup}), with training and testing on GSM8K, and Llama3.2-3B-Instruct as the base LRM
(\Cref{tab:ablation_qwen}).
The results of {\em Zero-shot} are taken from \cite{Extra-CoT}, while {\em DAPO} refers to our reproduction of DAPO \cite{DAPO} using the publicly available code and the hyperparameter values suggested by the authors. Note that {\em DAPO} in \Cref{tab:ablation_qwen} includes the length penalty 
(${\cal R}_{len}(\cdot)$)
defined in \Cref{eq.OverlongPunishment}.
The only  hyperparameter specific to \ours is the number of expansions $K$ (\Cref{alg.Tree}). If $K = 0$, then \ours reduces to DAPO. 
\Cref{tab:ablation_qwen} shows that, by varying the value of $K$ from 0 ({\em DAPO}) to 3, both AES and average length significantly improve, while accuracy with $K = 3$ is slightly higher than DAPO.
In all the other experiments of this paper, we use $K = 3$.

%%%%%%%%%%%%%%%%%%%%%%%%%%%%%%%%%%%%%%%%%%%%%%%%%%%%%%%%%%%%%%%%%%%%%%%%%%%%%%%%%%%%%%%%%%%%%%%%%%%%%%

\begin{table*}[!htb]
\caption{Ablation experiments using the ID protocol  and Llama3.2-3B-Instruct. Best results are shown in {\bf bold}, and second-best results are \underline{underlined}.}
\label{tab:ablation_qwen}
\centering
\small
\resizebox{0.6\textwidth}{!}{%
\begin{tabular}{lcc|ccc}
\toprule
\multirow{2}{*}{Methods} & \multirow{2}{*}{K} & \multirow{2}{*}{Len. Penalty}
& \multicolumn{3}{c}{GSM8K} \\
\cmidrule(lr){4-6}
& & 
& Ave. Len.$\downarrow$ & Acc$\uparrow$ & AES$\uparrow$ \\
\midrule

Zero-shot & -- & -- 
& 214 & 79.5 & 0.0 \\

DAPO & 0 & \checkmark 
& 200 & 85.6 & 0.3 \\

\midrule

\ours & 1 & \checkmark 
& 189 & 85.5 & 0.34 \\

\ours & 2 & \checkmark 
& 169 & 85.3 & 0.43 \\

\ours (full method) & 3 & \checkmark 
& \underline{151} & \underline{85.8} & \textbf{0.53} \\

w/o BFS & 3 & \checkmark 
& 204 & 85.6 & 0.28 \\

w/o Entropy & 3 & \checkmark 
& \textbf{145} & 82.7 & 0.44 \\

w/o ${\cal R}_{len}(\cdot)$ & 3 & \xmark 
& 167 & \textbf{86.0} & \underline{0.46} \\

\bottomrule
\end{tabular}}
\end{table*}

%%%%%%%%%%%%%%%%%%%%%%%%%%%%%%%%%%%%%%%%%%%%%%%%%%%%%%%%%%%%%%%%%%%%%%%%%%%%%%%%%%%%%%%%%%%%%%%%%%%%%%

\begin{table*}[!htb]
\caption{Model comparison with in-domain experiments on GSM8K.}
\label{tab:gsm8k_2}
\centering
\small
\resizebox{\textwidth}{!}{%
\begin{tabular}{ll|ccc|ccc|ccc|ccc|ccc|ccc}
\toprule
\multirow{2}{*}{Methods} & \multirow{2}{*}{Ratio}
& \multicolumn{3}{c|}{Qwen2.5-3B-Instruct}
& \multicolumn{3}{c|}{Qwen2.5-7B-Instruct}
& \multicolumn{3}{c|}{Qwen2.5-14B-Instruct}
& \multicolumn{3}{c|}{Llama3.2-3B-Instruct}
& \multicolumn{3}{c|}{LLaMA-3.1-8B-Instruct}
& \multicolumn{3}{c}{Average} \\
\cmidrule(lr){3-5}
\cmidrule(lr){6-8}
\cmidrule(lr){9-11}
\cmidrule(lr){12-14}
\cmidrule(lr){15-17}
\cmidrule(lr){18-20}
& & Ave. Len.$\downarrow$ & Acc$\uparrow$ & AES$\uparrow$
  & Ave. Len.$\downarrow$ & Acc$\uparrow$ & AES$\uparrow$
  & Ave. Len.$\downarrow$ & Acc$\uparrow$ & AES$\uparrow$
  & Ave. Len.$\downarrow$ & Acc$\uparrow$ & AES$\uparrow$
  & Ave. Len.$\downarrow$ & Acc$\uparrow$ & AES$\uparrow$
  & Ave. Len.$\downarrow$ & Acc$\uparrow$ & AES$\uparrow$ \\
\midrule

Zero-shot & --
& 315 & 83.7 & 0.00
& 297 & 91.4 & 0.00
& 313 & 93.1 & 0.00
& 214 & 79.5 & 0.00
& 213 & 86.2 & 0.00
& 270 & 86.8 & 0.00 \\
\midrule

\multirow{6}{*}{TokenSkip \cite{tokenskip}}
& 1.0
& 319 & 83.4 & -0.03
& 296 & 91.7 & 0.01
& 315 & 93.0 & -0.01
& 199 & 80.5 & 0.11
& 214 & 86.7 & 0.02
& 269 & 87.1 & 0.02 \\

& 0.9
& 263 & 83.2 & 0.13
& 255 & 91.1 & 0.12
& 269 & 93.3 & 0.15
& 182 & 78.0 & 0.05
& 198 & 86.1 & 0.06
& 233 & 86.3 & 0.10 \\

& 0.8
& 251 & 81.6 & 0.07
& 237 & 90.1 & 0.13
& 247 & 93.2 & 0.21
& 162 & 77.0 & 0.08
& 170 & 84.3 & 0.09
& 213 & 85.2 & 0.12 \\

& 0.7
& 233 & 80.1 & 0.04
& 217 & 89.9 & 0.18
& 219 & 93.4 & 0.31
& 153 & 74.1 & -0.05
& 150 & 82.5 & 0.08
& 194 & 84.0 & 0.11 \\

& 0.6
& 200 & 77.3 & -0.01
& \underline{178} & 87.9 & 0.20
& \underline{181} & 92.7 & 0.40
& 141 & 73.9 & -0.01
& 129 & 81.1 & 0.10
& 166 & 82.6 & 0.14 \\

& 0.5
& \textbf{171} & 74.4 & -0.09
& \textbf{151} & 86.0 & 0.19
& \textbf{157} & 91.4 & \underline{0.41}
& \underline{123} & 72.5 & -0.01
& \textbf{113} & 78.2 & 0.01
& \textbf{143} & 80.5 & 0.10 \\
\midrule

\multirow{6}{*}{Extra-CoT \cite{Extra-CoT}}
& 1.0
& -- & -- & --
& 298 & \underline{92.3} & 0.02
& -- & -- & --
& 196 & 81.5 & 0.15
& -- & -- & --
& 247 & 86.9 & 0.09 \\

& 0.9
& -- & -- & --
& 283 & 92.1 & 0.07
& -- & -- & --
& 181 & 79.3 & 0.14
& -- & -- & --
& 232 & 85.7 & 0.11 \\

& 0.8
& -- & -- & --
& 261 & 91.7 & 0.13
& -- & -- & --
& 163 & 79.2 & 0.21
& -- & -- & --
& 212 & 85.4 & 0.17 \\

& 0.7
& -- & -- & --
& 236 & 90.5 & 0.15
& -- & -- & --
& 145 & 78.5 & 0.25
& -- & -- & --
& 191 & 84.5 & 0.20 \\

& 0.6
& -- & -- & --
& 222 & 89.9 & 0.17
& -- & -- & --
& 132 & 77.0 & 0.22
& -- & -- & --
& 177 & 83.5 & 0.20 \\

& 0.5
& -- & -- & --
& 189 & 89.4 & \underline{0.25}
& -- & -- & --
& \textbf{115} & 73.4 & 0.07
& -- & -- & --
& \underline{152} & 81.4 & 0.16 \\
\midrule

Tree Of Thoughts \cite{treeThoughts} & --
& -- & -- & --
& -- & -- & --
& -- & -- & --
& -- & -- & --
& 385 & 74.0 & -1.51
& 385 & 74.0 & -1.51 \\

TreeRL & --
& 293 & 85.8 & 0.15
& 258 & 92.3 & 0.16
& 297 & 94.4 & 0.09
& 212 & 85.4 & 0.23
& 198 & 88.8 & 0.16
& 252 & 89.3 & 0.16 \\

DAPO \cite{DAPO} & --
& 250 & \textbf{87.4} & \underline{0.34}
& 235 & \textbf{92.8} & \underline{0.25}
& 244 & \underline{94.7} & 0.27
& 200 & \underline{85.6} & \underline{0.30}
& 194 & \textbf{89.6} & \underline{0.21}
& 225 & \textbf{90.0} & \underline{0.27} \\

\rowcolor{gray!30}
\ours & --
& \underline{198} & \underline{86.3} & \textbf{0.46}
& 200 & \textbf{92.8} & \textbf{0.37}
& 193 & \textbf{95.0} & \textbf{0.45}
& 151 & \textbf{85.8} & \textbf{0.53}
& \underline{125} & \underline{89.5} & \textbf{0.53}
& 173 & \underline{89.9} & \textbf{0.47} \\

\bottomrule
\end{tabular}}
\end{table*}

%%%%%%%%%%%%%%%%%%%%%%%%%%%%%%%%%%%%%%%%%%%%%%%%%%%%%%%%%%%%%%%%%%%%%%%%%%%%%%%%%%%%%%%%%%%%%%%%%%%%%%

\begin{table*}[t]
\centering
\footnotesize
\setlength{\tabcolsep}{5pt}
\caption{Model comparison with out-of-domain experiments and Qwen3-1.7B.}
\label{tab:main_table}
\resizebox{\textwidth}{!}{%
\begin{tabular}{lc|ccc|ccc|ccc|ccc|ccc}
\toprule
\multirow{2}{*}{Method} & \multirow{2}{*}{Ratio} &
\multicolumn{3}{c|}{\textbf{GSM8K}} &
\multicolumn{3}{c|}{\textbf{MATH-500}} &
\multicolumn{3}{c|}{\textbf{AMC2023}} &
\multicolumn{3}{c|}{\textbf{MMLU (STEM)}} &
\multicolumn{3}{c}{\textbf{Average}} \\
\cmidrule(lr){3-5}\cmidrule(lr){6-8}\cmidrule(lr){9-11}
\cmidrule(lr){12-14}\cmidrule(lr){15-17}
& & Ave. Len.\,$\downarrow$ & Acc.\,$\uparrow$ & AES\,$\uparrow$ &
Ave. Len.\,$\downarrow$ & Acc.\,$\uparrow$ & AES\,$\uparrow$ &
Ave. Len.\,$\downarrow$ & Acc.\,$\uparrow$ & AES\,$\uparrow$ &
Ave. Len.\,$\downarrow$ & Acc.\,$\uparrow$ & AES\,$\uparrow$ &
Ave. Len.\,$\downarrow$ & Acc.\,$\uparrow$ & AES\,$\uparrow$ \\
\midrule

Zero-shot & -- &
873 & 86.8 & 0.00 &
1675 & 64.2 & 0.00 &
2092 & 47.5 & 0.00 &
1081 & 74.1 & 0.00 &
1430 & 68.2 & 0.00 \\

\midrule
LC-Prompt \cite{lee2025how} & 0.6 &
763 & 86.2 & 0.09 &
1466 & 60.4 & -0.17 &
2015 & 45.0 & -0.23 &
- & - & - &
1415 & 63.9 & -0.10 \\

Truncation \cite{lee2025how} & 0.6 &
692 & 81.7 & -0.09 &
1271 & 49.8 & -0.88 &
1334 & 25.0 & -2.01 &
- & - & - &
1099 & 52.2 & -0.99 \\

\midrule
\multirow{6}{*}{TokenSkip \cite{tokenskip}}
& 1.0 &
916 & 83.7 & -0.23 &
1696 & 54.4 & -0.78 &
2333 & 37.5 & -1.17 &
1556 & 56.0 & -1.66 &
1625 & 57.9 & -0.96 \\
& 0.8 &
794 & 84.3 & -0.05 &
1527 & 53.0 & -0.78 &
2302 & 27.5 & -2.21 &
1481 & 54.5 & -1.69 &
1526 & 54.8 & -1.18 \\
& 0.6 &
770 & 78.5 & -0.36 &
1517 & 47.4 & -1.21 &
2077 & 20.0 & -2.89 &
1310 & 47.5 & -2.01 &
1418 & 48.4 & -1.62 \\
& 0.4 &
516 & 74.2 & -0.32 &
1259 & 34.4 & -2.07 &
1743 & 12.5 & -3.52 &
877 & 29.8 & -2.80 &
1099 & 37.7 & -2.18 \\
& 0.2 &
\underline{273} & 59.1 & -0.91 &
660 & 23.4 & -2.57 &
911 & 10.0 & -3.38 &
\underline{446} & 18.1 & -3.19 &
572 & 27.7 & -2.51 \\
& \texttt{<POLICY>} &
473 & 70.0 & -0.51 &
1182 & 35.8 & -1.92 &
1437 & 12.5 & -3.37 &
1035 & 32.6 & -2.76 &
1032 & 37.7 & -2.14 \\

\midrule
\multirow{6}{*}{Extra-CoT \cite{Extra-CoT}}
& 1.0 &
902 & 86.7 & -0.04 &
1698 & 64.0 & -0.03 &
2153 & 42.5 & -0.56 &
1252 & 70.5 & -0.40 &
1501 & 65.9 & -0.26 \\
& 0.8 &
807 & 86.7 & 0.07 &
1520 & 64.2 & 0.09 &
1845 & 42.5 & -0.41 &
1267 & 71.3 & -0.36 &
1360 & 66.2 & -0.15 \\
& 0.6 &
641 & 85.5 & 0.19 &
1304 & 59.6 & -0.14 &
1624 & 40.0 & -0.57 &
1012 & 60.6 & -0.85 &
1145 & 61.4 & -0.34 \\
& 0.4 &
469 & 82.3 & 0.20 &
920 & 54.2 & -0.33 &
945 & 25.0 & -1.82 &
730 & 61.3 & -0.54 &
766 & 55.7 & -0.62 \\
& 0.2 &
303 & 80.2 & 0.27 &
\underline{481} & 47.8 & -0.56 &
\underline{782} & 17.5 & -2.53 &
458 & 54.3 & -0.76 &
\underline{506} & 49.9 & -0.90 \\
& \texttt{<POLICY>} &
569 & 86.3 & 0.32 &
1312 & 63.0 & 0.12 &
1515 & 42.5 & -0.25 &
1026 & 67.5 & -0.39 &
1106 & 64.8 & -0.05 \\

\midrule
Thinkless (DeGRPO) \cite{fang2026thinkless} & -- &
356 & 85.5 & 0.52 &
888 & 63.6 & 0.42 &
1369 & 50.0 & 0.50 &
- & - & - &
871 & 66.4 & 0.47 \\

Extra-CoT (CHRPO) \cite{Extra-CoT} & \texttt{<POLICY>} &
\textbf{210} & 85.8 & \textbf{0.70} &
\textbf{452} & 64.8 & 0.76 &
\textbf{675} & 50.0 & 0.84 &
\textbf{330} & 67.1 & 0.22 &
\textbf{417} & 66.9 & 0.63 \\

\midrule
TreeRL \cite{treerl} & -- &
786$\pm$9 & 87.1$\pm$0.4 & 0.11$\pm$0.03 &
1618$\pm$42 & 76.2$\pm$0.8 & 0.59$\pm$0.05 &
2050$\pm$95 & 59.2$\pm$3.6 & 0.76$\pm$0.23 &
932$\pm$9 & 76.8$\pm$0.4 & 0.25$\pm$0.03 &
1347$\pm$28 & 74.8$\pm$0.9 & 0.43$\pm$0.06 \\

DAPO \cite{DAPO} & -- &
606$\pm$7 & \underline{89.2$\pm$0.4} & 0.39$\pm$0.02 &
1307$\pm$31 & \underline{80.7$\pm$0.8} & \underline{1.00$\pm$0.04} &
2011$\pm$92 & \underline{60.7$\pm$3.4} & \underline{0.87$\pm$0.22} &
663$\pm$7 & \underline{79.0$\pm$0.4} & \underline{0.58$\pm$0.02} &
1147$\pm$24 & \underline{77.4$\pm$0.8} & \underline{0.71$\pm$0.05} \\

\rowcolor{gray!30}
\ours & -- &
437$\pm$2 & \textbf{89.7$\pm$0.4} & \underline{0.60$\pm$0.01} &
968$\pm$16 & \textbf{81.6$\pm$0.7} & \textbf{1.24$\pm$0.04} &
1558$\pm$76 & \textbf{64.7$\pm$4.4} & \textbf{1.34$\pm$0.26} &
459$\pm$2 & \textbf{79.3$\pm$0.4} & \textbf{0.79$\pm$0.02} &
856$\pm$19 & \textbf{78.8$\pm$0.9} & \textbf{0.99$\pm$0.05} \\

\bottomrule
\end{tabular}%
}
\end{table*}

%%%%%%%%%%%%%%%%%%%%%%%%%%%%%%%%%%%%%%%%%%%%%%%%%%%%%%%%%%%%%%%%%%%%%%%%%%%%%%%%%%%%%%%%%%%%%%%%%%%%%%

In \Cref{tab:ablation_qwen}, {\em w/o BFS} refers to our tree search algorithm without BFS. Specifically,
in this case, the backtracking point is {\em not} selected using the current best solution but only using the entropy. In more detail,
in {\em w/o BFS}, we remove line 5 from \Cref{alg.Tree} and we replace line 6 with:

\begin{equation}
    o_{j,t^*} =   \arg \max_{o_{j,t} \in {\cal T}, o_{j,t} \notin B} H(o_{j,t}).
\end{equation}

We use {\em w/o BFS} to evaluate the contribution of a BFS strategy in which only the current best solution is expanded. Since  in {\em w/o BFS}  we expand the highest entropy node selected from all the current complete solutions (i.e., from all the paths from the root to a leaf),
it  can be considered similar to a breadth-first  search strategy.
The results in \Cref{tab:ablation_qwen} show that {\em w/o BFS} can boost the accuracy with respect to {\em Zero-shot}, 
but the reduction in the average number of generated tokens is much less pronounced compared to our full method (\ours with $K = 3$),
which  shows  the effectiveness of a  BFS strategy in reducing overthinking. We believe this is because our BFS strategy biases the LRM exploration towards progressively shorter solutions, thereby self-generating a training dataset of  concise CoTs (\Cref{sec.Introduction}).

The next entry in \Cref{tab:ablation_qwen} is {\em w/o Entropy}, in which
we select the backtracking point randomly instead of using entropy. Specifically, {\em w/o Entropy} corresponds to replacing step 6 in \Cref{alg.Tree} with a random selection of $o_{j,t^*}$ from all tokens of $o_j$, leaving everything else unchanged ({\em including the BFS strategy}, i.e., the selection
of $o_j$ using \Cref{eq.best-solution} in step 5). The results in \Cref{tab:ablation_qwen} show that both accuracy and AES are much lower in {\em w/o Entropy} than in our full method, despite the average length being shorter.
This shows that, although the BFS strategy in {\em w/o Entropy} leads to an effective CoT shortening, using entropy to select backtracking points is crucial to increasing accuracy. This is likely due to the fact that, since these high entropy tokens are the points where LRMs make the main decisions \cite{beyond-80-20},  our branching approach allows the model to better semantically differentiate the solutions explored at training time and, consequently, better generalize at inference time
(\Cref{sec.Discussion}).
%Comparing {\em w/o BFS} with {\em w/o Entropy},  it becomes clear that using a shortest solution-guided search is crucial to shorten CoTs, while %entropy is needed to increase accuracy.

Finally, in {\em w/o  ${\cal R}_{len}(\cdot)$}, we remove the  length penalty of DAPO (\Cref{eq.OverlongPunishment}) 
from \Cref{eq.DAPO-tot-reward}
and we use a (leaf) reward based only on the sequence correctness ($R_i = 1$ if $v(o_i,a^*) = 1$ and $R_i = -1$ if $v(o_i,a^*) = 0$), keeping all the rest of our method  unchanged.
Overall, the results of {\em w/o  ${\cal R}_{len}(\cdot)$} are slightly worse than our full method, however they are much better than {\em DAPO}, {\em in which ${\cal R}_{len}(\cdot)$ is used}, which shows that the benefits of \ours depend only marginally from the length penalty.

\subsection{Main results}
\label{sec.MainExp}

In \Cref{tab:gsm8k_2} we use the ID experimental protocol (\Cref{sec.ExpSetup}) to compare \ours with other methods.  
The results of Zero-shot (i.e.,  the base models)  and TokenSkip are taken from \cite{tokenskip}.
Specifically, TokenSkip with compression ratio = 1 corresponds to SFT (without CoT shortening and using the TokenSkip prompt).
Moreover,
the results of
Extra-CoT  are taken from  \cite{Extra-CoT}, while  
 the results of Tree of Thought \cite{treeThoughts} (\Cref{sec.Related}) are taken from \cite{mcdonald-etal-2025-afford}, and
 DAPO \cite{DAPO} is our reproduction.
In \Cref{tab:main_table} we use the OOD experimental protocol  (\Cref{sec.ExpSetup}).
The results of  Zero-shot,
LC-Prompt \cite{lee2025how}, Truncation \cite{lee2025how}, 
 TokenSkip \cite{tokenskip},  Extra-CoT \cite{Extra-CoT}, Thinkless (DeGRPO) \cite{fang2026thinkless} and  Extra-CoT with RL (CHRPO) \cite{Extra-CoT}
are  taken from \cite{Extra-CoT}.
Additionally, in the same table we include 
our reproduction of DAPO and 
TreeRL \cite{treerl} using the publicly available code and the hyperparameter values suggested by the corresponding authors.
For those methods for which we have access to the trained parameters (\ours,  DAPO and TreeRL),
we repeated the testing phase  using 16 different seeds, and we report in \Cref{tab:main_table} the mean and the standard deviation\footnote{The other results in \Cref{tab:main_table} are taken from \cite{Extra-CoT}, where the authors have not released the parameters of the tested methods, while the ID protocol used in \Cref{tab:gsm8k_2,tab:Qwen3-4B-Base}
 is based on greedy decoding, thus we cannot repeat testing using multiple seeds.}.

\Cref{tab:gsm8k_2} and \Cref{tab:main_table} show that  BFS-PO   
{\em consistently  achieves the best AES score}, with both protocols and with all models and datasets, 
and often by a wide margin over other methods, 
with the sole exception in \Cref{tab:main_table}, for GSM8K, where \ours achieves the second-best AES score.
Furthermore, in all experiments, BFS-PO consistently achieves either the best or the second-best result in terms of accuracy. Specifically, the average accuracy in \Cref{tab:gsm8k_2} is nearly equal to that of DAPO, while in \Cref{tab:main_table} it outperforms DAPO by about one and a half points.
When compared with Zero-shot, \ours significantly decreases the CoT length  {\em and} increases the accuracy. For instance, considering the results averaged over the four  datasets in  \Cref{tab:main_table}, the mean CoT length of \ours is 574 tokens 
shorter than Zero-shot ($-40.1\%$), while  accuracy  improved by $10.6$ points.
Comparing the average length of the CoTs produced by \ours with those of TreeRL and DAPO in \Cref{tab:main_table}, \ours shortens the generated sequences by $36.5\%$ and $25.4\%$, respectively.
TokenSkip and Extra-CoT can compress more than \ours, 
but only at the cost of a drastic reduction in accuracy compared to Zero-shot.

Finally, in \Cref{tab:Qwen3-4B-Base} we use the ID experimental protocol with a much larger  budget of 4096 tokens
and Qwen3-4B-Base. The latter is 
a ``Base'' model—that is, a model lacking a post-training phase—and, consequently,  it is characterized by  a relatively high level of output uncertainty.
The  results in \Cref{tab:Qwen3-4B-Base}
show that, even when using a model with a high average token entropy, \ours can largely decrease the CoT length 
($-34.9\%$ with respect to Zero-shot)
while simultaneously increasing the accuracy. We believe that this is due to the fact that, as shown in \cite{beyond-80-20}, token entropy is a signal of logical forking even for models without post-training, and is therefore a fairly universal indicator of those points in an LLM's reasoning trajectory where the model makes decisions and chooses the main paths.

\begin{table}[t]
\centering
\footnotesize
\setlength{\tabcolsep}{6pt}
\caption{Model comparison with  Qwen3-4B-Base.}
\label{tab:Qwen3-4B-Base}
\begin{tabular}{lccc}
\toprule
& \multicolumn{3}{c}{\textbf{GSM8K}} \\
\cmidrule(lr){2-4}
Method & Ave. Len. $\downarrow$ & Acc. $\uparrow$ & AES $\uparrow$ \\
\midrule
Zero-shot & 301 & 91.0 & 0.00 \\
TreeRL    & 304 & 92.8 & 0.05 \\
DAPO      & \underline{298} & \underline{93.1} & \underline{0.08} \\
\rowcolor{gray!30}

\ours     & \textbf{196} & \textbf{93.6} & \textbf{0.43} \\
\bottomrule
\end{tabular}
\end{table}

Overall, the results
reported in \Cref{tab:gsm8k_2,tab:main_table,tab:Qwen3-4B-Base}
show that \ours 
significantly reduces the average CoT length with respect to Zero-shot while still achieving an accuracy equal to or better than DAPO.

\section{Conclusions}
\label{sec.Conclusion}

We proposed \ours, an RL algorithm for LRM in which the solution space is explored using a Best-First Search. Specifically,
we avoid the use of a costly PRM by selecting the
best solution from the  complete  solutions of the current search tree, where
a standard verifier can be applied and the path length measured. We then use the highest entropy tokens for
backtracking and tree expansion, following recent literature
showing the importance of these tokens in LRM decisions. Using different benchmarks and models, we showed
that BFS-PO can {\em simultaneously} increase accuracy {\em and}
shorten the average CoT length, in stark contrast
to state-of-the-art CoT shortening methods that typically
sacrifice accuracy for length reduction.

{\bf Limitations.} In our experiments we strictly followed the experimental protocols proposed in \cite{Extra-CoT,tokenskip}, and 
we used LRMs with 1.7B to 14B parameters, which is one of the most common settings for CoT shortening methods. While extending the experiments using larger models would be interesting, training larger models using RL is very computationally expensive
and beyond our computational resources, and we believe that the experimental protocols we have adopted can be used as a basis for comparison by a greater number of competitors.

\subsubsection*{Acknowledgments}

This work was supported by the PNRR project ``Italian Strengthening of Esfri RI Resilience (ITSERR)'', funded by the European Union – NextGenerationEU (CUP B53C22001770006) and
 by the EU Horizon project ELIAS (No.
101120237). 
We acknowledge EuroHPC JU for awarding the project ID EHPC-AIF-2026SC01-043 access to Leonardo Booster
hosted by CINECA,
and  the CINECA award
under the ISCRA initiative
for the availability of high-performance computing resources and support.

\bibliographystyle{ieeetr}  %{plainnat}
\bibliography{bibliography}

%%%%%%%%%%%%%%%%%%%%%%%%%%%%%%%%%%%%%%%%%%%%%%%%%%%%%%%%%%%%

\newpage
\appendix
\raggedbottom

\section{GRPO and DAPO}
\label{app.GRPO}

The GRPO objective is \cite{GRPO}:

\begin{equation}
\label{eq.GRPO}
\begin{split}
& {\cal J}_{GRPO} = \mathbb{E}_{(q,a^*) \sim {\cal D}, \{o_i\}_{i=1}^G \sim \pi_{\theta_{old}}(\cdot|q)} \\
& \left[ \frac{1}{G} \sum_{i=1}^G \frac{1}{|o_i|} \sum_{t=1}^{|o_i|} 
\left( \min (r_{i,t}(\theta) \hat{A}_{i,t}, clip(r_{i,t}(\theta), 1 - \epsilon, 1 + \epsilon) \hat{A}_{i,t} - \beta D_{KL}(\pi_{\theta} || \pi_{ref}))
\right) 
\right], 
\end{split}
\end{equation}

\noindent
where:

\begin{equation}
\label{eq.advantage-GRPO-app}
\hat{A}_{i,t} = \frac{R_i - mean(\{ R_i \}_{i=1}^G)}{std(\{ R_i \}_{i=1}^G)}, \: \:
r_{i,t}(\theta) = \frac{\pi_{\theta}(o_{i,t}|q, o_{i,<t})}{\pi_{\theta_{old}}(o_{i,t}|q, o_{i,<t})}.
\end{equation}

In \Cref{eq.GRPO}, the Kullback-Leibler divergence  $D_{KL}(\pi_{\theta} || \pi_{ref})$  is used to 
regularize
the difference between the online policy ($\pi_{\theta}$) and the frozen reference
policy ($\pi_{ref}$), where the latter is usually the base LRM model before the RL fine-tuning starts.
$\beta$ is an hyperparameter which weights this KL penalty, while $\epsilon$ is the clipping-related
hyperparameter introduced in PPO for stabilizing training \cite{PPO}.

In DAPO \cite{DAPO}, the KL penalty is removed, while the clipping operation is  bounded below by  $\epsilon_l$ and above by $\epsilon_u$. Moreover, Yu et al. \cite{DAPO}
 observe that, if all the $G$ answers are correct or if they are all wrong, then the advantage is 0. 
Therefore, they resample those groups of outputs which do not contain  at least one correct and one incorrect answer (see \Cref{sec.Preliminaries}). Finally, Yu et al. \cite{DAPO}
show that the normalization term $\frac{1}{|o_i|}$ in \Cref{eq.GRPO}, 
which divides the objective function value of a given sample $o_i$ by its length, penalizes long correct answers ($v(o_i,a^*) = 1$) but also has the opposite effect, i.e. decreases the penalty for long incorrect answers ($v(o_i,a^*) = 0$) \cite{fatemi2025concise,colar}. The authors  empirically show that this leads to longer CoTs.
Therefore, they replace this term by a group-based normalization formulated as $\frac{1}{\sum_{i=1}^G |o_i|}$:

\begin{equation}
\label{eq.DAPO-app}
\begin{split}
& {\cal J}_{DAPO} (\theta) = \mathbb{E}_{(q,a^*) \sim {\cal D}, \{o_i\}_{i=1}^G \sim \pi_{\theta_{old}}(\cdot|q)} \\
& \left[ \frac{1}{\sum_{i=1}^G |o_i|} \sum_{i=1}^G \sum_{t=1}^{|o_i|} 
\left( min (r_{i,t}(\theta) \hat{A}_{i,t}, clip(r_{i,t}(\theta), 1 - \epsilon_l, 1 + \epsilon_h) \hat{A}_{i,t} 
\right) 
\right] \\
& s.t. \: \: 0 < |\{ o_i | v(o_i,a^*) = 1 \}| <  G.
\end{split}
\end{equation}

As mentioned in \Cref{sec.Preliminaries}, the constraint $0 < |\{ o_i | v(o_i,a^*) = 1 \}| <  G$ 
is used to ensure a non-zero advantage, and it is implemented by resampling the training problems which violate this constraint (\Cref{sec.Preliminaries}). Finally, very long sequences are penalized using Soft Overlong Punishment, defined as:

 \begin{equation}
 \label{eq.OverlongPunishment-app}
{\cal R}_{len}(o) = 
\begin{cases}
    0 & \text{$|o| \leq L_{max} - L_{cache}$}\\
    \frac{(L_{max} - L_{cache}) - |o|}{L_{cache}} & \text{$L_{max} - L_{cache} < |o| \leq L_{max}$}\\
    -1 & \text{$L_{max} < |o|$,}\\
  \end{cases}
\end{equation}
 
\noindent
and the final reward is computed using:

 \begin{equation}
 \label{eq.DAPO-tot-reward-app}
R_i = 
\begin{cases}
    1 + {\cal R}_{len}(o_i) & \text{$v(o_i,a^*) = 1$}\\
    -1 + {\cal R}_{len}(o_i) & \text{$v(o_i,a^*) = 0$.}\\
  \end{cases}
\end{equation}

In all our experiments with DAPO we adopt the  hyperparameter values 
suggested in  \cite{DAPO}: $\epsilon_l = 0.2$, $\epsilon_h = 0.28$, 
$L_{cache} = \frac{1}{5} L_{budget}$ and $L_{max} = L_{budget} - L_{cache}$, where
 $L_{budget}$ is the  token budget, i.e., the maximum number of tokens which can be generated by the LRM (and  depends on the specific computational resources).
For a fair comparison, in all experiments with \ours, 
{\em we  use the same hyperparameter values} ($\epsilon_l = 0.2$, $\epsilon_h = 0.28$, 
$L_{cache} = \frac{1}{5} L_{budget}$,  $L_{max} = L_{budget} - L_{cache}$), which are kept fixed without any variation.

\section{Experimental details}
\label{app.Details}

Our implementation is based on the TRL framework \cite{TRL}. We fine-tune all models using LoRA
\cite{lora}, with rank $r=8$ and $\alpha=16$. We keep the optimization and
training hyperparameters fixed across all experiments. Specifically, we use AdamW
with learning rate $1\mathrm{e}{-6}$ and 10 warmup steps.
We use a generation batch size of 2048, except for Qwen2.5-14B-Instruct in \Cref{tab:gsm8k_2} and Qwen3-1.7B in \Cref{tab:main_table}, for which we use 4096. In all experiments, we use a temperature of 1.0 and top-p of 1.0 for rollout generation, and a training batch size of 512 with a minibatch size of 32, corresponding to 16 training steps per training batch. We monitor the AES
score during training and, for each method, we select the checkpoint corresponding to its highest AES score. For Qwen3-1.7B, which is evaluated on multiple datasets (\Cref{tab:main_table}), we monitor AES on MATH-500 (only) and we use the selected checkpoint for testing on all datasets. As mentioned in \Cref{sec.Preliminaries} and App. \ref{app.GRPO}, for both \ours and DAPO 
we adopt the hyperparameter values used in DAPO \cite{DAPO}: $\epsilon_h=0.28$
and $\epsilon_l=0.2$ for clipping, and $L_{cache} = \frac{1}{5} L_{budget}$,
$L_{max} = L_{budget} - L_{cache}$ for ${\cal R}_{len}(\cdot)$. Following the TokenSkip \cite{tokenskip} protocol, when training on GSM8K, at inference time we use greedy decoding,  and we set the maximum token budget $L_{\text{budget}}=512$, except for the additional experiments with Qwen3-4B-Base (\Cref{tab:Qwen3-4B-Base}), for which we use $L_{\text{budget}}=4096$. In the experiments of \Cref{tab:main_table}, following the Extra-CoT protocol, we use $L_{\text{budget}}=4096$, and at inference time a temperature of 0.7 and top-p of 0.9. As mentioned in \Cref{sec.Discussion}, for each base LRM we use the same training time budget for all methods. Specifically, for the experiments reported in \Cref{tab:gsm8k_2}, we use a training time budget of 24 hours for Qwen2.5-3B-Instruct and Llama3.2-3B-Instruct, and 72 hours for Qwen2.5-7B-Instruct and Llama3.1-8B-Instruct, with 4 NVIDIA A100 GPUs in all four cases. For Qwen2.5-14B-Instruct (\Cref{tab:gsm8k_2}) and Qwen3-1.7B (\Cref{tab:main_table}), we use a training time budget of 168 hours on 16 NVIDIA A100 GPUs. Finally, for Qwen3-4B-Base (\Cref{tab:Qwen3-4B-Base}), we use a training time budget of 24 hours on 8 NVIDIA A100 GPUs.

\section{Metrics}
\label{app.AES}

{\bf Average Length.}
Following \cite{Extra-CoT}, in the OOD setup, when computing the length of an output sequence $|o|$,  the token count  is calculated using only the CoT tokens contained within the 
$\texttt{<think>}$ ...$\texttt{</think>}$  block. On the other hand, 
following \cite{tokenskip,Extra-CoT},
 in the ID protocol, $|o|$ includes also the  final answer tokens\footnote{This distinction in the adopted protocols is presumably due to the fact that the  {\em instruct} models used in the ID scenario  do not separate the answer from the other tokens using special formatting.}.

{\bf Accuracy Efficiency Score (AES).}
The AES metric, proposed in \cite{O1-Pruner}, 
 combines the improvement in accuracy and the reduction
in computational cost compared to the baseline model. It weights and sums the differences between the evaluated
model and the baseline, taking into account both the average length of the reasoning chains and their accuracy. In our case, the baseline corresponds to the zero-shot performance of the model.
The terms $\Delta\text{Length}$ and $\Delta\text{Acc}$ represent these relative changes, and they are computed as follows \cite{O1-Pruner}:

\begin{equation}
\Delta\text{Length} = \frac{\text{Length}_{\text{baseline}} - \text{Length}_{\text{model}}}{\text{Length}_{\text{baseline}}}, \quad \Delta\text{Acc} = \frac{\text{Acc}_{\text{model}} - \text{Acc}_{\text{baseline}}}{\text{Acc}_{\text{baseline}}}.
\end{equation}

Using these terms, the AES score is defined as:

\begin{equation}
\text{AES} = 
\begin{cases} 
\alpha \cdot \Delta\text{Length} + \beta \cdot |\Delta\text{Acc}|, & \text{if } \Delta\text{Acc} \geq 0 \\ 
\alpha \cdot \Delta\text{Length} - \gamma \cdot |\Delta\text{Acc}|, & \text{if } \Delta\text{Acc} < 0. 
\end{cases}
\end{equation}

The coefficients $\alpha$, $\beta$ and $\gamma$ are used to weight the different components of the score. In our experiments we use the default hyperparameters values suggested in \cite{O1-Pruner}, which are $\alpha = 1, \beta = 3, \text{ and } \gamma = 5$.

\section{Training curves}
\label{app.TrainingCurves}

\begin{figure}[!htbp]
    
    \centering
    \includegraphics[
        width=0.9\linewidth,
        height=0.9\textheight,
        keepaspectratio
    ]{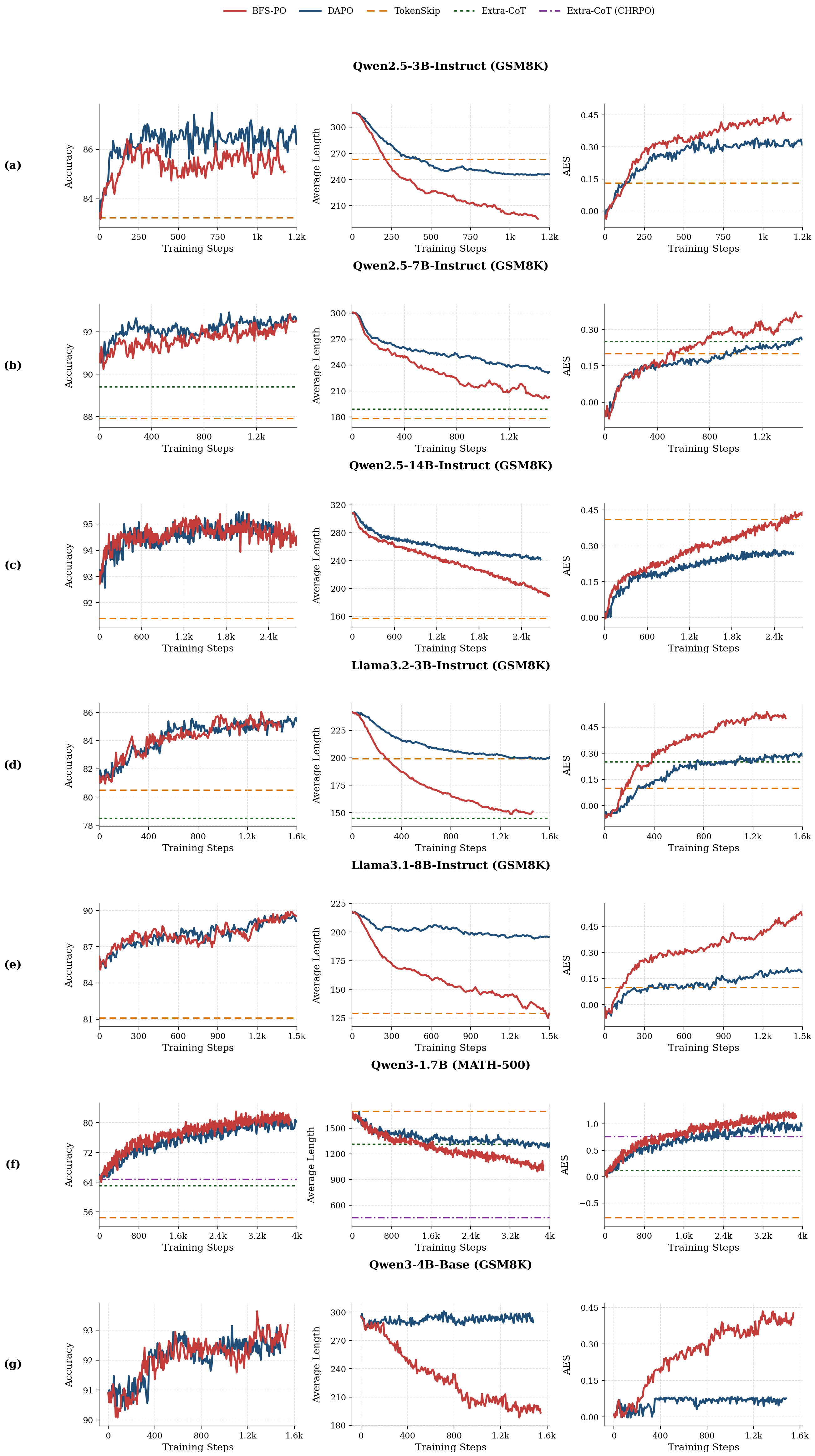}
        \caption{Training curves of \ours and DAPO for the experiments in \Cref{tab:gsm8k_2,tab:main_table,tab:Qwen3-4B-Base}. Evaluation during training is performed on GSM8K for all models except Qwen3-1.7B, which is evaluated on MATH-500 (see App. \ref{app.Details}). Additionally,  for TokenSkip, Extra-CoT and Extra-CoT (CHRPO) we show with dashed lines the  values corresponding to the best AES results reported in \Cref{tab:gsm8k_2,tab:main_table} and taken from \cite{tokenskip,Extra-CoT}.}
        \label{fig.training_curves_gsm8k}
\end{figure}

\Cref{fig.training_curves_gsm8k} shows the training curves of \ours and DAPO for both the ID and OOD experimental protocols (\Cref{sec.Experiments}).
Specifically,  rows (a)-(e) correspond to the ID experiments on GSM8K in \Cref{tab:gsm8k_2},
row (f) refers to the OOD experiments with Qwen3-1.7B in \Cref{tab:main_table}, and 
 row (g) corresponds to the additional ID experiments with Qwen3-4B-Base in \Cref{tab:Qwen3-4B-Base}. Note that, in all graphs, each training step on the horizontal axis corresponds to 16 parameter updates (App. \ref{app.Details}).
As mentioned in \Cref{sec.Discussion} and App. \ref{app.Details}, both  DAPO and \ours are trained with the same training time budget  and the two methods complete approximately the same number of training steps. Independently of the model, \ours has always a higher AES than DAPO during the entire  training. Moreover, \ours largely outperforms both TokenSkip and Extra-CoT, whose best AES values are significantly lower than the AES achieved by \ours. These graphs  also show that the average length of \ours is consistently lower than that of DAPO throughout training, which confirms that our method learns shorter reasoning chains from the early stages of training. Finally, the accuracy curves of \ours and DAPO stay close to each other, showing that the reduction in reasoning length does not come at the cost of a loss in accuracy.

\section{Training time comparison}
\label{app.Training-time-comparison}

In \Cref{tab:wall-clock}, we compare the  wall-clock training time of \ours with  DAPO and TreeRL.
Specifically,
``Training Time to Best'' refers to the time needed to reach the highest AES score, which is the criterion  used to select the best checkpoint (\Cref{app.Details}). Moreover, ``Tab. $X$” refers
 to the experiments in the $X$-th table of this paper.
 \Cref{tab:wall-clock} shows that the three methods require comparable training times on average. Specifically, the average ``Training Time to Best'' is 76.3 hours for \ours, 72.9 hours for DAPO, and 76.9 hours for TreeRL. Compared with DAPO, \ours requires about $4.7\%$ more time on average, while it requires about $0.8\%$ less time than TreeRL. Finally, since we keep the hardware, batch size, token budget, and memory allocation fixed across the three methods, \ours, DAPO and TreeRL use the same amount of VRAM in each experiment. The VRAM and wall-clock time results show no evidence of memory or communication bottlenecks for \ours.

\begin{table*}[t]
\centering
\footnotesize
\setlength{\tabcolsep}{5pt}
\caption{Wall-clock training time and VRAM usage of \ours, DAPO, and TreeRL.}
\label{tab:wall-clock}
\resizebox{0.66\textwidth}{!}{%
\begin{tabular}{llcccc}
\toprule
 & Model & Method &
\shortstack{Training time\\budget (h)} &
\shortstack{Training time\\to best (h)} &
\shortstack{VRAM usage\\(GB)} \\
\midrule

\multirow{15}{*}{\Cref{tab:gsm8k_2}}
& \multirow{3}{*}{Qwen2.5-3B-Instruct}
& \ours & \multirow{3}{*}{24} & 22 & \multirow{3}{*}{40} \\
& & DAPO   & & 22 & \\
& & TreeRL & & 23 & \\
\cmidrule(lr){2-6}

& \multirow{3}{*}{Qwen2.5-7B-Instruct}
& \ours & \multirow{3}{*}{72} & 72 & \multirow{3}{*}{42} \\
& & DAPO   & & 71 & \\
& & TreeRL & & 71 & \\
\cmidrule(lr){2-6}

& \multirow{3}{*}{Qwen2.5-14B-Instruct}
& \ours & \multirow{3}{*}{168} & 166 & \multirow{3}{*}{62} \\
& & DAPO   & & 158 & \\
& & TreeRL & & 166 & \\
\cmidrule(lr){2-6}

& \multirow{3}{*}{Llama3.2-3B-Instruct}
& \ours & \multirow{3}{*}{24} & 21 & \multirow{3}{*}{40} \\
& & DAPO   & & 22 & \\
& & TreeRL & & 23 & \\
\cmidrule(lr){2-6}

& \multirow{3}{*}{Llama3.1-8B-Instruct}
& \ours & \multirow{3}{*}{72} & 71 & \multirow{3}{*}{42} \\
& & DAPO   & & 69 & \\
& & TreeRL & & 70 & \\

\midrule

\multirow{3}{*}{\Cref{tab:main_table}}
& \multirow{3}{*}{Qwen3-1.7B}
& \ours & \multirow{3}{*}{168} & 162 & \multirow{3}{*}{61} \\
& & DAPO   & & 150 & \\
& & TreeRL & & 164 & \\
\midrule

\multirow{3}{*}{\Cref{tab:Qwen3-4B-Base}}
& \multirow{3}{*}{Qwen3-4B-Base}
& \ours & \multirow{3}{*}{24} & 20 & \multirow{3}{*}{52} \\
& & DAPO   & & 18 & \\
& & TreeRL & & 21 & \\

\bottomrule
\end{tabular}%
}
\end{table*}

\section{Qualitative analysis}
\label{app.Qualitative}

\subsection{Solution space exploration during training}
\label{app.Qual-Solution-space}

In this section, we show how \ours progressively generates shorter correct CoTs during the tree construction
using the example in \Cref{fig.qualitative}. Specifically, given a question, the LRM initially generates $G$ answers 
(line 2 of \Cref{alg.Tree}). 
Line 5 of \Cref{alg.Tree}
 selects the shortest correct answer $o_1$, which, in this example, has $277$ tokens (see \Cref{fig.qualitative}).
 Line 6 selects  the first backtracking token $w_1^b$, which corresponds to \colorbox{highlight}{pieces}. 
 Then, line 7  samples $G$ partial output sequences starting from $w_1^b$ (i.e., using the text
 ``\texttt{\detokenize{Let}} ...  \texttt{\detokenize{pieces}}''
 as a suffix to the question). In the second iteration, \ours selects  the new shortest correct solution,  which, in this example,
  is $o_2$, that differs from $o_1$ starting from the token \colorbox{continuation}{her}. 
  In $o_2$, the new selected backtracking point ($w_2^b$) is \colorbox{highlight}{pies}, which is used for the second expansion.
  In the third iteration, the new shortest correct output is $o_3$, with only $198$ tokens. Note that $o_3$  is $28\%$ shorter than  $o_1$.

\begin{center}
\end{center}
\begin{tcolorbox}[
    colback=boxbg,
    colframe=black!85,
    title=\textbf{Exploration of the solution space in \ours},
    fonttitle=\small\bfseries,
    arc=0pt,
    left=5pt,
    right=5pt]
\scriptsize

\textbf{Question ($q$):}
Grace baked 4 whole pumpkin pies. She sold 1 whole pumpkin pie and gave 1 whole pumpkin pie to her friend. The remaining whole pumpkin pies were each sliced into 6 pieces. Her family ate 2/3 pieces. How many pieces of pumpkin pie were left?

\medskip
\textbf{Answer: 4}

\vspace{0.8em}
\hrule
\vspace{0.8em}

% ===================== A1 =====================
\textbf{$o_1$ (277 tokens):} \\
\begingroup
\ttfamily \hyphenchar\font=-1
\detokenize{Let's break down the problem step-by-step:\n\n1. Grace baked 4 whole pumpkin pies.\n2. She sold 1 whole pumpkin pie.\n3. She gave 1 whole pumpkin pie to her friend.\n4. Therefore, the number of pumpkin pies remaining is:\n\\[\n4 - 1 - 1 = 2 \\text{ whole pumpkin pies}\n\\]\n\n5. Each of the remaining 2 whole pumpkin pies was sliced into 6 pieces:\n\\[\n2 \\text{ pies} \\times 6 \\text{ pieces per pie} = 12 \\text{ pieces}\n\\]\n\n6. Her family ate \\(\\frac{2}{3}\\) of the total pieces of pie. To find out how many pieces were eaten:\n\\[\n\\frac{2}{3} \\text{ of } 12 \\text{ pieces} = \\frac{2}{3} \\times 12 = 8 \\text{ pieces}\n\\]\n\n7. To find out how many } \highlighttoken{highlight}{pieces} \detokenize{ } \highlighttoken{continuation}{are} \detokenize{ left after the family ate:\n\\[\n12 \\text{ pieces} - 8 \\text{ pieces} = 4 \\text{ pieces}\n\\]\n\nTherefore, the number of pieces of pumpkin pie left is:\n\\[\n\\boxed{4}\n\\]}
\endgroup

\vspace{1.5em}
\hrule
\vspace{0.8em}

% ===================== A2 =====================
\textbf{$o_2$ (241 tokens):} \\
\begingroup
\ttfamily \hyphenchar\font=-1
\detokenize{Let's break down the problem step-by-step:\n\n1. Grace baked 4 whole pumpkin pies.\n2. She sold 1 whole pumpkin pie.\n3. She gave 1 whole pumpkin pie to her friend.\n4. Therefore, the number of pumpkin pies remaining is:\n\\[\n4 - 1 - 1 = 2 \\text{ whole pumpkin pies}\n\\]\n\n5. Each of the remaining 2 whole pumpkin pies was sliced into 6 pieces:\n\\[\n2 \\text{ pies}\\times 6 \\text{ pieces per pie} = 12 \\text{ pieces}\n\\]\n\n6. Her family ate \\(\\frac{2}{3}\\) of the total pieces of pie. To find out how many} \highlighttoken{highlight}{pieces} \detokenize{ } \highlighttoken{continuation}{her} \detokenize{ family ate:\n\\[\n\\frac{2}{3} \\times 12 = 8 \\text{ pieces}\n\\]\n\n7. To find the number of pieces left:\n\\[\n  12 - 8 = 4 \\text{ pieces}\n\\]\n\nTherefore, the number of pieces of pumpkin pie that were left is \\(\\boxed{4}\\).}
\endgroup

\vspace{1.5em}
\hrule
\vspace{0.8em}

% ===================== A2_2 =====================
\textbf{$o_2$ (241 tokens):} \\
\begingroup
\ttfamily \hyphenchar\font=-1
\detokenize{Let's break down the problem step-by-step:\n\n1. Grace baked 4 whole pumpkin pies.\n2. She sold 1 whole pumpkin pie.\n3. She gave 1 whole pumpkin pie to her friend.\n4. Therefore, the number of pumpkin} \highlighttoken{highlight}{pies} \detokenize{ } \highlighttoken{continuation}{remaining} \detokenize{ is:\n\\[\n4 - 1 - 1 = 2 \\text{ whole pumpkin pies}\n\\]\n\n5. Each of the remaining 2 whole pumpkin pies was sliced into 6 pieces:\n\\[\n2 \\text{ pies}\\times 6 \\text{ pieces per pie} = 12 \\text{ pieces}\n\\]\n\n6. Her family ate \\(\\frac{2}{3}\\) of the total pieces of pie. To find out how many pieces her family ate:\n\\[\n\\frac{2}{3} \\times 12 = 8 \\text{ pieces}\n\\]\n\n7. To find the number of pieces left:\n\\[\n  12 - 8 = 4 \\text{ pieces}\n\\]\n\nTherefore, the number of pieces of pumpkin pie that were left is \\(\\boxed{4}\\).}
\endgroup

\vspace{1.5em}
\hrule
\vspace{0.8em}

% ===================== A3 =====================
\textbf{$o_3$ (198 tokens):} \\
\begingroup
\ttfamily \hyphenchar\font=-1
\detokenize{Let's break down the problem step-by-step:\n\n1. Grace baked 4 whole pumpkin pies.\n2. She sold 1 whole pumpkin pie.\n3. She gave 1 whole pumpkin pie to her friend.\n4. Therefore, the number of whole pumpkin} \highlighttoken{highlight}{pies} \detokenize{ } \highlighttoken{continuation}{left} \detokenize{ is \\(4 - 1 - 1 = 2\\).\n5. Each of the remaining 2 whole pumpkin pies is sliced into 6 pieces.\n6. Thus, the total number of pie pieces is \\(2 \\times 6 = 12\\) pieces.\n7. Her family ate \\(\\frac{2}{3}\\) of the total pieces.\n8. The number of pieces her family ate is \\(\\frac{2}{3} \\times 12 = 8\\) pieces.\n9. Therefore, the number of pieces left is \\(12 - 8 = 4\\).\n\nPutting the final answer within the \\boxed{}:\n\\[\n\\boxed{4}\n\\]}
\endgroup

\end{tcolorbox}
\captionsetup{hypcap=false} % 
\captionof{figure}{A qualitative example of the Best-First Search strategy  using Qwen2.5-3B-Instruct during  training on  GSM8K.}
\label{fig.qualitative}

%\nopagebreak % Evita che la didascalia finisca in una pagina diversa dall'ultima parte della box

\subsection{Inference-time comparison with the base model}
\label{app.Qual-Inference-time}

% In this section, we show a 
% qualitative analysis which compares CoT generated by the Zero-shot model with CoT generated by BSF-PO for the same questions. To make the comparison easier, we used the same color to highlight the reasoning steps that are semantically equivalent in both responses. This analysis shows that BSF-PO leads to shortening LRM outputs by removing non-essential reasoning details, such as, for example, the units of measurement and other verbose aspects of the response that do not affect its logical structure, which is exactly the goal that an overthinking shortening method should pursue. 

\Cref{fig.qualitative_2} compares \ours with the base model (Zero-shot) {\em at inference time}, and  shows that \ours generates shorter reasoning chains without skipping essential reasoning steps. Specifically, both the base model  and \ours generate reasoning sequences which can be partitioned into  five logical  steps,  highlighted in \Cref{fig.qualitative_2} using the same color. The main difference concerns the way the reasoning step is expressed. In the \blocktag{strongred}{first block}, the base model introduces the first subproblem with: "\quotett{Calculate the amount of dog food used in the first 180 days}", while \ours uses a shorter sentence: "\quotett{First 180 days}". However, both intermediate reasoning chains address the same subproblem and reach the same result. In the \blocktag{stepblue}{second block}, the base model generates: "\quotett{Subtract the first 180 days from the total days in a year}", while \ours expresses the same operation more directly as: "\quotett{Remaining days = 365 - 180 = 185 days}". The main compression appears in the \blocktag{stepgreen}{third block} and \blocktag{steporange}{fourth block}. The base model first introduces the second-period computation in the \blocktag{stepgreen}{third block} with: "\quotett{Therefore, the total amount of dog food used in the remaining 185 days is}" and then combines the two quantities in the \blocktag{steporange}{fourth block} with: "\quotett{Add the dog food used in the first 180 days and the remaining 185 days}". \ours, instead, keeps in the \blocktag{stepgreen}{third block} only the premise: "\quotett{For the first year, we only need to consider the first 185 days}", and then performs both the multiplication and the aggregation in the \blocktag{steporange}{fourth block} through: "\quotett{Next 185 days: 185 * 2 = 370 cups}" and "\quotett{Total for the first year = 180 + 370 = 550 cups}". This comparison shows that \ours does not skip the computation of 370 cups. It simply moves it to the next block and expresses it more compactly. In the \blocktag{steppurple}{fifth block}, the base model generates: "\quotett{Divide the total amount of dog food by the number of cups per bag}", while \ours generates only the final calculation: "\quotett{Number of bags = 550 / 110 = 5 bags}". 

\begin{tcolorbox}[ colback=white, colframe=black!85, title=\textbf{Reasoning Chain Comparison: Zero-shot vs BFS-PO (Qwen2.5-3B-Instruct)}, fonttitle=\small\bfseries, arc=0pt, left=0pt, right=0pt] 
\scriptsize \textbf{Question ($q$):} \texttt{\detokenize{Cecilia just bought a new puppy. According to her veterinarian, she has to feed the puppy 1 cup of dog food every day for the first 180 days. Then she has to feed the puppy 2 cups of dog food every day for the rest of its life. If one bag of dog food contains 110 cups, how many bags of dog food will Cecilia use in the first year?}}
\medskip 
\\
\textbf{Answer: 5} 
\vspace{0.5em} 
\hrule 
\vspace{0.5em} 

\textbf{Zero-shot (504 tokens):} \\ 
\begingroup 
\ttfamily \hyphenchar\font=-1 
\detokenize{To determine how many bags of dog food Cecilia will use in the first year, we need to break down the problem into two parts: the first 180 days and the rest of the puppy's life.\n\n} 
\endgroup

\begin{highlightblock}{strongred}
\detokenize{1. **Calculate the amount of dog food used in the first 180 days:**\n - The puppy eats 1 cup of dog food per day for the first 180 days.\n - Therefore, the total amount of dog food used in the first 180 days is:\n \\[\n 1 \\text{ cup/day} \\times 180 \\text{ days} = 180 \\text{ cups}\n \\]\n\n}
\end{highlightblock}

\begingroup 
\ttfamily \hyphenchar\font=-1 
\endgroup

\begin{highlightblock}{stepblue}
\detokenize{2. **Determine the number of days remaining in the first year:**\n - A year has 365 days.\n - Subtract the first 180 days from the total days in a year:\n \\[\n 365 \\text{ days} - 180 \\text{ days} = 185 \\text{ days}\n \\]\n\n}
\end{highlightblock}

\begin{highlightblock}{stepgreen}
\detokenize{3. **Calculate the amount of dog food used in the remaining 185 days:**\n - After the first 180 days, the puppy eats 2 cups of dog food per day.\n - Therefore, the total amount of dog food used in the remaining 185 days is:\n \\[\n 2 \\text{ cups/day} \\times 185 \\text{ days} = 370 \\text{ cups}\n \\]\n\n}
\end{highlightblock}

\begin{highlightblock}{steporange}
\detokenize{4. **Calculate the total amount of dog food used in the first year:**\n - Add the dog food used in the first 180 days and the remaining 185 days:\n \\[\n 180 \\text{ cups} + 370 \\text{ cups} = 550 \\text{ cups}\n \\]\n\n}
\end{highlightblock}

\begin{highlightblock}{steppurple}
\detokenize{5. **Determine the number of bags of dog food needed:**\n - Each bag contains 110 cups of dog food.\n - Divide the total amount of dog food by the number of cups per bag:\n \\[\n \\frac{550 \\text{ cups}}{110 \\text{ cups/bag}} = 5 \\text{ bags}\n \\]\n\n}
\end{highlightblock}

\begingroup 
\ttfamily \hyphenchar\font=-1 
\detokenize{Thus, Cecilia will use \\boxed{5} bags of dog food in the first year.} 
\endgroup 

\vspace{0.5em} 
\hrule 
\vspace{0.5em}

\textbf{BFS-PO (320 tokens):} \\ 
\begingroup 
\ttfamily \hyphenchar\font=-1 
\detokenize{To determine how many bags of dog food Cecilia will use in the first year, we need to calculate the total amount of dog food required for the first 180 days and then for the rest of the puppy's life within the first year.\n\n} 
\endgroup

\begin{highlightblock}{strongred}
\detokenize{1. **First 180 days:**\n - The puppy needs 1 cup of dog food per day.\n - Total for 180 days = 180 cups.\n\n}
\end{highlightblock}

\begin{highlightblock}{stepblue}
\detokenize{2. **Remaining days in the first year:**\n - A year has 365 days.\n - Remaining days = 365 - 180 = 185 days.\n\n}
\end{highlightblock}

\begin{highlightblock}{stepgreen}
\detokenize{3. **After the first year:**\n - The puppy will need 2 cups of dog food per day for the rest of its life.\n - For the first year, we only need to consider the first 185 days.\n\n}
\end{highlightblock}

\begin{highlightblock}{steporange}
\detokenize{4. **Total dog food needed for the first year:**\n - First 180 days: 180 cups.\n - Next 185 days: 185 * 2 = 370 cups.\n - Total for the first year = 180 + 370 = 550 cups.\n\n}
\end{highlightblock}

\begin{highlightblock}{steppurple}
\detokenize{5. **Number of bags required:**\n - Each bag contains 110 cups.\n - Number of bags = 550 / 110 = 5 bags.\n\n}
\end{highlightblock}

\begingroup 
\ttfamily \hyphenchar\font=-1 
\detokenize{Thus, Cecilia will use \\boxed{5} bags of dog food in the first year.} 
\endgroup 
\end{tcolorbox} 
    \captionsetup{hypcap=false} % Disabilita l'opzione che genera il warning
    \captionof{figure}{A qualitative comparison of the generated reasoning chains for a GSM8K problem using Qwen2.5-3B-Instruct, which shows the inference-time differences between Zero-shot and \ours.}
    \label{fig.qualitative_2}
%%%%%%%%%%%%%%%%%%%%%%%%%%%%%%%%%%%%%%%%%%%%%%%%%%%%%%%%%%%%

\clearpage
\section*{NeurIPS Paper Checklist}

\begin{enumerate}

\item {\bf Claims}
    \item[] Question: Do the main claims made in the abstract and introduction accurately reflect the paper's contributions and scope?
    \item[] Answer: \answerYes{}
    \item[] Justification: In the abstract and introduction we claim that (1) \ours can alleviate the overthinking problem and (2) we sketch the main methodological aspects of our proposal. The first claim is empirically validated in \Cref{sec.Experiments}, where \ours is compared to different SOTA baselines, while the second is detailed in \Cref{sec.Method}.
    \item[] Guidelines:
    \begin{itemize}
        \item The answer \answerNA{} means that the abstract and introduction do not include the claims made in the paper.
        \item The abstract and/or introduction should clearly state the claims made, including the contributions made in the paper and important assumptions and limitations. A \answerNo{} or \answerNA{} answer to this question will not be perceived well by the reviewers. 
        \item The claims made should match theoretical and experimental results, and reflect how much the results can be expected to generalize to other settings. 
        \item It is fine to include aspirational goals as motivation as long as it is clear that these goals are not attained by the paper. 
    \end{itemize}

\item {\bf Limitations}
    \item[] Question: Does the paper discuss the limitations of the work performed by the authors?
    \item[] Answer: \answerYes{}.
    \item[] Justification: We discussed the limitations of our experiments in \Cref{sec.Conclusion}, in an ad hoc paragraph called "Limitations", in which we say that, for computational reasons, we cannot use models larger than 14B  parameters.
    \item[] Guidelines:
    \begin{itemize}
        \item The answer \answerNA{} means that the paper has no limitation while the answer \answerNo{} means that the paper has limitations, but those are not discussed in the paper. 
        \item The authors are encouraged to create a separate ``Limitations'' section in their paper.
        \item The paper should point out any strong assumptions and how robust the results are to violations of these assumptions (e.g., independence assumptions, noiseless settings, model well-specification, asymptotic approximations only holding locally). The authors should reflect on how these assumptions might be violated in practice and what the implications would be.
        \item The authors should reflect on the scope of the claims made, e.g., if the approach was only tested on a few datasets or with a few runs. In general, empirical results often depend on implicit assumptions, which should be articulated.
        \item The authors should reflect on the factors that influence the performance of the approach. For example, a facial recognition algorithm may perform poorly when image resolution is low or images are taken in low lighting. Or a speech-to-text system might not be used reliably to provide closed captions for online lectures because it fails to handle technical jargon.
        \item The authors should discuss the computational efficiency of the proposed algorithms and how they scale with dataset size.
        \item If applicable, the authors should discuss possible limitations of their approach to address problems of privacy and fairness.
        \item While the authors might fear that complete honesty about limitations might be used by reviewers as grounds for rejection, a worse outcome might be that reviewers discover limitations that aren't acknowledged in the paper. The authors should use their best judgment and recognize that individual actions in favor of transparency play an important role in developing norms that preserve the integrity of the community. Reviewers will be specifically instructed to not penalize honesty concerning limitations.
    \end{itemize}

\item {\bf Theory assumptions and proofs}
    \item[] Question: For each theoretical result, does the paper provide the full set of assumptions and a complete (and correct) proof?
    \item[] Answer: \answerNA{}.
    \item[] Justification: This paper does not include theoretical results.
    \item[] Guidelines:
    \begin{itemize}
        \item The answer \answerNA{} means that the paper does not include theoretical results. 
        \item All the theorems, formulas, and proofs in the paper should be numbered and cross-referenced.
        \item All assumptions should be clearly stated or referenced in the statement of any theorems.
        \item The proofs can either appear in the main paper or the supplemental material, but if they appear in the supplemental material, the authors are encouraged to provide a short proof sketch to provide intuition. 
        \item Inversely, any informal proof provided in the core of the paper should be complemented by formal proofs provided in appendix or supplemental material.
        \item Theorems and Lemmas that the proof relies upon should be properly referenced. 
    \end{itemize}

    \item {\bf Experimental result reproducibility}
    \item[] Question: Does the paper fully disclose all the information needed to reproduce the main experimental results of the paper to the extent that it affects the main claims and/or conclusions of the paper (regardless of whether the code and data are provided or not)?
    \item[] Answer: \answerYes{}.
    \item[] Justification: We provide our code, the network parameters and the evaluation procedure with the Supplementary Material. Moreover, our method is explained in \Cref{sec.Method} and the experimental protocols (including the public datasets we used) in \Cref{sec.ExpSetup}. Finally, the Appendix sections \ref{app.GRPO}, \ref{app.Details} and \ref{app.AES} provide the details and all the hyperparameter values, thus our method and our experiments are fully-reproducible.
    \item[] Guidelines:
    \begin{itemize}
        \item The answer \answerNA{} means that the paper does not include experiments.
        \item If the paper includes experiments, a \answerNo{} answer to this question will not be perceived well by the reviewers: Making the paper reproducible is important, regardless of whether the code and data are provided or not.
        \item If the contribution is a dataset and\slash or model, the authors should describe the steps taken to make their results reproducible or verifiable. 
        \item Depending on the contribution, reproducibility can be accomplished in various ways. For example, if the contribution is a novel architecture, describing the architecture fully might suffice, or if the contribution is a specific model and empirical evaluation, it may be necessary to either make it possible for others to replicate the model with the same dataset, or provide access to the model. In general. releasing code and data is often one good way to accomplish this, but reproducibility can also be provided via detailed instructions for how to replicate the results, access to a hosted model (e.g., in the case of a large language model), releasing of a model checkpoint, or other means that are appropriate to the research performed.
        \item While NeurIPS does not require releasing code, the conference does require all submissions to provide some reasonable avenue for reproducibility, which may depend on the nature of the contribution. For example
        \begin{enumerate}
            \item If the contribution is primarily a new algorithm, the paper should make it clear how to reproduce that algorithm.
            \item If the contribution is primarily a new model architecture, the paper should describe the architecture clearly and fully.
            \item If the contribution is a new model (e.g., a large language model), then there should either be a way to access this model for reproducing the results or a way to reproduce the model (e.g., with an open-source dataset or instructions for how to construct the dataset).
            \item We recognize that reproducibility may be tricky in some cases, in which case authors are welcome to describe the particular way they provide for reproducibility. In the case of closed-source models, it may be that access to the model is limited in some way (e.g., to registered users), but it should be possible for other researchers to have some path to reproducing or verifying the results.
        \end{enumerate}
    \end{itemize}

\item {\bf Open access to data and code}
    \item[] Question: Does the paper provide open access to the data and code, with sufficient instructions to faithfully reproduce the main experimental results, as described in supplemental material?
    \item[] Answer:  \answerYes{}.
    \item[] Justification: In the Supplementary Material we provided our code and all the instructions to execute it and reproduce the main results (our code will be published after the acceptance of this paper). Moreover, we do not use private data but only public datasets. 
    \item[] Guidelines:
    \begin{itemize}
        \item The answer \answerNA{} means that paper does not include experiments requiring code.
        \item Please see the NeurIPS code and data submission guidelines (\url{https://neurips.cc/public/guides/CodeSubmissionPolicy}) for more details.
        \item While we encourage the release of code and data, we understand that this might not be possible, so \answerNo{} is an acceptable answer. Papers cannot be rejected simply for not including code, unless this is central to the contribution (e.g., for a new open-source benchmark).
        \item The instructions should contain the exact command and environment needed to run to reproduce the results. See the NeurIPS code and data submission guidelines (\url{https://neurips.cc/public/guides/CodeSubmissionPolicy}) for more details.
        \item The authors should provide instructions on data access and preparation, including how to access the raw data, preprocessed data, intermediate data, and generated data, etc.
        \item The authors should provide scripts to reproduce all experimental results for the new proposed method and baselines. If only a subset of experiments are reproducible, they should state which ones are omitted from the script and why.
        \item At submission time, to preserve anonymity, the authors should release anonymized versions (if applicable).
        \item Providing as much information as possible in supplemental material (appended to the paper) is recommended, but including URLs to data and code is permitted.
    \end{itemize}

\item {\bf Experimental setting/details}
    \item[] Question: Does the paper specify all the training and test details (e.g., data splits, hyperparameters, how they were chosen, type of optimizer) necessary to understand the results?
    \item[] Answer:  \answerYes{}.
    \item[] Justification: We followed two training/testing protocols, respectively taken from \cite{Extra-CoT} and \cite{tokenskip}. These are  clearly explained in  \Cref{sec.ExpSetup}, with additional details in \Cref{app.Details} and \Cref{app.AES}. In  \Cref{app.GRPO} and \Cref{app.Details} we provide all the hyperparameter values, type of optimizer used and all the the training and testing details.
    \item[] Guidelines:
    \begin{itemize}
        \item The answer \answerNA{} means that the paper does not include experiments.
        \item The experimental setting should be presented in the core of the paper to a level of detail that is necessary to appreciate the results and make sense of them.
        \item The full details can be provided either with the code, in appendix, or as supplemental material.
    \end{itemize}

\item {\bf Experiment statistical significance}
    \item[] Question: Does the paper report error bars suitably and correctly defined or other appropriate information about the statistical significance of the experiments?
    \item[] Answer:  \answerNo{}.
    \item[] Justification: Unfortunately, error bars or other types of statistical measures which need to repeat the training phase  more than once are computationally infeasible for the LRM domain, especially when using RL-based algorithms. As far as we know, no paper reports these kinds of statistics. 
    However, testing is much more computationally efficient, thus we repeated the experiments in \Cref{tab:main_table} using 16 different seeds and we reported the means and the standard deviations. Please, note that the experiments in \Cref{tab:gsm8k_2,tab:Qwen3-4B-Base} are based on the greedy decoding protocol proposed in \cite{tokenskip}, thus we cannot repeat testing using multiple seeds.
    Importantly, our experiments are based on multiple datasets, and   \Cref{tab:gsm8k_2,tab:main_table,tab:Qwen3-4B-Base}
    show that \ours consistently beats SOTA on all the datasets and settings, demonstrating the scientific robustness of our results. 
    \item[] Guidelines:
    \begin{itemize}
        \item The answer \answerNA{} means that the paper does not include experiments.
        \item The authors should answer \answerYes{} if the results are accompanied by error bars, confidence intervals, or statistical significance tests, at least for the experiments that support the main claims of the paper.
        \item The factors of variability that the error bars are capturing should be clearly stated (for example, train/test split, initialization, random drawing of some parameter, or overall run with given experimental conditions).
        \item The method for calculating the error bars should be explained (closed form formula, call to a library function, bootstrap, etc.)
        \item The assumptions made should be given (e.g., Normally distributed errors).
        \item It should be clear whether the error bar is the standard deviation or the standard error of the mean.
        \item It is OK to report 1-sigma error bars, but one should state it. The authors should preferably report a 2-sigma error bar than state that they have a 96\% CI, if the hypothesis of Normality of errors is not verified.
        \item For asymmetric distributions, the authors should be careful not to show in tables or figures symmetric error bars that would yield results that are out of range (e.g., negative error rates).
        \item If error bars are reported in tables or plots, the authors should explain in the text how they were calculated and reference the corresponding figures or tables in the text.
    \end{itemize}

\item {\bf Experiments compute resources}
    \item[] Question: For each experiment, does the paper provide sufficient information on the computer resources (type of compute workers, memory, time of execution) needed to reproduce the experiments?
    \item[] Answer:  \answerYes{}.
    \item[] Justification: The computer resources for all our experiments are provided in \Cref{app.Details}.
    \item[] Guidelines:
    \begin{itemize}
        \item The answer \answerNA{} means that the paper does not include experiments.
        \item The paper should indicate the type of compute workers CPU or GPU, internal cluster, or cloud provider, including relevant memory and storage.
        \item The paper should provide the amount of compute required for each of the individual experimental runs as well as estimate the total compute. 
        \item The paper should disclose whether the full research project required more compute than the experiments reported in the paper (e.g., preliminary or failed experiments that didn't make it into the paper). 
    \end{itemize}
    
\item {\bf Code of ethics}
    \item[] Question: Does the research conducted in the paper conform, in every respect, with the NeurIPS Code of Ethics \url{https://neurips.cc/public/EthicsGuidelines}?
    \item[] Answer:  \answerYes{}.
    \item[] Justification: Our research respects all the details of NeurIPS Code of Ethics \url{https://neurips.cc/public/EthicsGuidelines}.
    \item[] Guidelines:
    \begin{itemize}
        \item The answer \answerNA{} means that the authors have not reviewed the NeurIPS Code of Ethics.
        \item If the authors answer \answerNo, they should explain the special circumstances that require a deviation from the Code of Ethics.
        \item The authors should make sure to preserve anonymity (e.g., if there is a special consideration due to laws or regulations in their jurisdiction).
    \end{itemize}

\item {\bf Broader impacts}
    \item[] Question: Does the paper discuss both potential positive societal impacts and negative societal impacts of the work performed?
    \item[] Answer:  \answerNo{}.
    \item[] Justification: Our contribution is entirely methodological and application agnostic: it provides a generic method for limiting overthinking in LRMs, without targeting any particular domain or use case, and therefore does not raise any specific social or ethical concerns that need to be highlighted in this context.
    \item[] Guidelines:
    \begin{itemize}
        \item The answer \answerNA{} means that there is no societal impact of the work performed.
        \item If the authors answer \answerNA{} or \answerNo, they should explain why their work has no societal impact or why the paper does not address societal impact.
        \item Examples of negative societal impacts include potential malicious or unintended uses (e.g., disinformation, generating fake profiles, surveillance), fairness considerations (e.g., deployment of technologies that could make decisions that unfairly impact specific groups), privacy considerations, and security considerations.
        \item The conference expects that many papers will be foundational research and not tied to particular applications, let alone deployments. However, if there is a direct path to any negative applications, the authors should point it out. For example, it is legitimate to point out that an improvement in the quality of generative models could be used to generate Deepfakes for disinformation. On the other hand, it is not needed to point out that a generic algorithm for optimizing neural networks could enable people to train models that generate Deepfakes faster.
        \item The authors should consider possible harms that could arise when the technology is being used as intended and functioning correctly, harms that could arise when the technology is being used as intended but gives incorrect results, and harms following from (intentional or unintentional) misuse of the technology.
        \item If there are negative societal impacts, the authors could also discuss possible mitigation strategies (e.g., gated release of models, providing defenses in addition to attacks, mechanisms for monitoring misuse, mechanisms to monitor how a system learns from feedback over time, improving the efficiency and accessibility of ML).
    \end{itemize}
    
\item {\bf Safeguards}
    \item[] Question: Does the paper describe safeguards that have been put in place for responsible release of data or models that have a high risk for misuse (e.g., pre-trained language models, image generators, or scraped datasets)?
    \item[] Answer:  \answerNA{}.
    \item[] Justification: We do not release any dataset or pre-trained model, but just a procedure to fine-tune an existing, pre-trained LRM.
    \item[] Guidelines:
    \begin{itemize}
        \item The answer \answerNA{} means that the paper poses no such risks.
        \item Released models that have a high risk for misuse or dual-use should be released with necessary safeguards to allow for controlled use of the model, for example by requiring that users adhere to usage guidelines or restrictions to access the model or implementing safety filters. 
        \item Datasets that have been scraped from the Internet could pose safety risks. The authors should describe how they avoided releasing unsafe images.
        \item We recognize that providing effective safeguards is challenging, and many papers do not require this, but we encourage authors to take this into account and make a best faith effort.
    \end{itemize}

\item {\bf Licenses for existing assets}
    \item[] Question: Are the creators or original owners of assets (e.g., code, data, models), used in the paper, properly credited and are the license and terms of use explicitly mentioned and properly respected?
    \item[] Answer:  \answerYes{}.
    \item[] Justification: We use only public datasets and public codes. We have cited all works, datasets, codes and assets that are mentioned in the paper.
    \item[] Guidelines:
    \begin{itemize}
        \item The answer \answerNA{} means that the paper does not use existing assets.
        \item The authors should cite the original paper that produced the code package or dataset.
        \item The authors should state which version of the asset is used and, if possible, include a URL.
        \item The name of the license (e.g., CC-BY 4.0) should be included for each asset.
        \item For scraped data from a particular source (e.g., website), the copyright and terms of service of that source should be provided.
        \item If assets are released, the license, copyright information, and terms of use in the package should be provided. For popular datasets, \url{paperswithcode.com/datasets} has curated licenses for some datasets. Their licensing guide can help determine the license of a dataset.
        \item For existing datasets that are re-packaged, both the original license and the license of the derived asset (if it has changed) should be provided.
        \item If this information is not available online, the authors are encouraged to reach out to the asset's creators.
    \end{itemize}

\item {\bf New assets}
    \item[] Question: Are new assets introduced in the paper well documented and is the documentation provided alongside the assets?
    \item[] Answer:  \answerNA{}.
    \item[] Justification: No new assets are released in this paper.
    \item[] Guidelines:
    \begin{itemize}
        \item The answer \answerNA{} means that the paper does not release new assets.
        \item Researchers should communicate the details of the dataset\slash code\slash model as part of their submissions via structured templates. This includes details about training, license, limitations, etc. 
        \item The paper should discuss whether and how consent was obtained from people whose asset is used.
        \item At submission time, remember to anonymize your assets (if applicable). You can either create an anonymized URL or include an anonymized zip file.
    \end{itemize}

\item {\bf Crowdsourcing and research with human subjects}
    \item[] Question: For crowdsourcing experiments and research with human subjects, does the paper include the full text of instructions given to participants and screenshots, if applicable, as well as details about compensation (if any)? 
    \item[] Answer:  \answerNA{}.
    \item[] Justification: This article does not contain any crowdsourcing experiments or research with human subjects.
    \item[] Guidelines:
    \begin{itemize}
        \item The answer \answerNA{} means that the paper does not involve crowdsourcing nor research with human subjects.
        \item Including this information in the supplemental material is fine, but if the main contribution of the paper involves human subjects, then as much detail as possible should be included in the main paper. 
        \item According to the NeurIPS Code of Ethics, workers involved in data collection, curation, or other labor should be paid at least the minimum wage in the country of the data collector. 
    \end{itemize}

\item {\bf Institutional review board (IRB) approvals or equivalent for research with human subjects}
    \item[] Question: Does the paper describe potential risks incurred by study participants, whether such risks were disclosed to the subjects, and whether Institutional Review Board (IRB) approvals (or an equivalent approval/review based on the requirements of your country or institution) were obtained?
    \item[] Answer:  \answerNA{}.
    \item[] Justification: This article does not contain any  participants.
    \item[] Guidelines:
    \begin{itemize}
        \item The answer \answerNA{} means that the paper does not involve crowdsourcing nor research with human subjects.
        \item Depending on the country in which research is conducted, IRB approval (or equivalent) may be required for any human subjects research. If you obtained IRB approval, you should clearly state this in the paper. 
        \item We recognize that the procedures for this may vary significantly between institutions and locations, and we expect authors to adhere to the NeurIPS Code of Ethics and the guidelines for their institution. 
        \item For initial submissions, do not include any information that would break anonymity (if applicable), such as the institution conducting the review.
    \end{itemize}

\item {\bf Declaration of LLM usage}
    \item[] Question: Does the paper describe the usage of LLMs if it is an important, original, or non-standard component of the core methods in this research? Note that if the LLM is used only for writing, editing, or formatting purposes and does \emph{not} impact the core methodology, scientific rigor, or originality of the research, declaration is not required.
    %this research? 
    \item[] Answer:  \answerYes{}.
    \item[] Justification: The method we propose in this article is an LLM fine-tuning approach, thus  the entire article is based on the detailed description of this method and how it is used with standard, publicly available LLMs.
    \item[] Guidelines:
    \begin{itemize}
        \item The answer \answerNA{} means that the core method development in this research does not involve LLMs as any important, original, or non-standard components.
        \item Please refer to our LLM policy in the NeurIPS handbook for what should or should not be described.
    \end{itemize}

\end{enumerate}

\end{document}